\documentclass[11pt]{article}

\usepackage[final]{acl}

\usepackage{times}
\usepackage{latexsym}

\usepackage[T1]{fontenc}

\usepackage[utf8]{inputenc}

\usepackage{microtype}

\usepackage{inconsolata}

\usepackage{graphicx}

\usepackage{subcaption}
\usepackage{booktabs}

\usepackage{amsmath}
\usepackage{amssymb}
\usepackage{mathtools}
\usepackage{amsthm}

\usepackage[capitalize,noabbrev]{cleveref}

\usepackage{amsmath,amsfonts,bm}

\def\eqref#1{equation~\ref{#1}}

\def\1{\bm{1}}

\DeclareMathAlphabet{\mathsfit}{\encodingdefault}{\sfdefault}{m}{sl}
\SetMathAlphabet{\mathsfit}{bold}{\encodingdefault}{\sfdefault}{bx}{n}

\usepackage{caption}  
\usepackage{listings} 
\usepackage{hyperref}
\usepackage{url}
\usepackage{booktabs}   
\usepackage{tabularx}
\usepackage{tikz}
\usepackage[textwidth=3cm]{todonotes} 
\usepackage{subcaption}  
\usepackage{graphicx}
\usepackage[export]{adjustbox} 
\usepackage[table]{xcolor} 
\usepackage{soul}
\definecolor{lightgrayrule}{gray}{0.7} 
\definecolor{checkgreen}{HTML}{90E681}  

\usepackage{pdflscape}
\usepackage{makecell}
\usepackage{enumitem}

\usepackage{longtable}
\usepackage{verbatim}
\usepackage{multirow}
\newcommand{\model}[1]{\textit{#1}}
\newcommand{\token}[1]{\texttt{#1}}
\usepackage{float}

\usepackage{fancyvrb}
\usepackage{cleveref} 
  
\definecolor{codebg}{HTML}{F5F7FA}        
\lstdefinestyle{plaintextfile}{
    basicstyle=\ttfamily\small,   
    numbers=left,                 
    numberstyle=\tiny,            
    stepnumber=1,                 
    numbersep=5pt,                
    showstringspaces=false,       
    breaklines=true,              
    breakatwhitespace=false,      
    frame=single,                 
    backgroundcolor=\color{codebg}, 
    moredelim=**[is][\color{green!60!black}\bfseries]{@@}{@@}, 
    moredelim=**[is][{\color{red}}]{§§}{§§},               
    breaklines=true,
    breakindent=0pt,   
    prebreak=,         
    postbreak=,        
}

\usepackage{etoolbox}
\makeatletter
\patchcmd{\section}{\@afterheading}{}{}{}
\makeatother

\newcommand{\samplecount}{$685,000$}
\newcommand{\repositorylink}{\href{https://github.com/muelphil/llm-uncertainty-bench}{https://github.com/muelphil/llm-uncertainty-bench}}
\newcommand{\frameworklink}{\href{https://github.com/Helmholtz-AI-Matter/async-graph-bench}{https://github.com/Helmholtz-AI-Matter/async-graph-bench}}

\title{\textbf{Semantic Calibration Prevails Where Token Confidence Fails: Benchmarking Long-Form Scientific QA}}

\author{
  \textbf{Philip Müller\textsuperscript{1}}\quad
  \textbf{Nicholas Popovič\textsuperscript{2,3}}\quad
  \textbf{Michael Färber\textsuperscript{2,3}}\quad
  \textbf{Peter Steinbach\textsuperscript{1}}
  \\
  \textsuperscript{1}Helmholtz-Zentrum Dresden-Rossendorf
  \quad
  \textsuperscript{2}Technische Universität Dresden
  \quad
  \textsuperscript{3}ScaDS.AI
  \\
  \small{
    \texttt{p.mueller@hzdr.de}
    \quad
    \texttt{nicholas.popovic@tu-dresden.de}
    \quad
    \texttt{michael.faerber@tu-dresden.de}
    \quad
    \texttt{p.steinbach@hzdr.de}
  }
}

\begin{document}
\maketitle
\begin{abstract}
Reliable uncertainty quantification (UQ) is essential for safe deployment of large language models (LLMs) in scientific question answering, where long-form outputs exceed practical human verification at scale. We introduce the first large-scale benchmark for UQ calibration in long-form, reasoning-demanding scientific QA, evaluating four UQ methods on \samplecount{} responses across up to 20 LLMs and seven datasets, supported by an extensible open-source framework whose shared-generation design enables reproducible cross-method comparisons. Instruction tuning is shown to associate with systematic token probability polarization, collapsing confidence distributions and undermining the reliability of token-level uncertainty signals. Reasoning model families diverge: some reproduce this polarization while others actively mitigate it, a pattern that clusters by provider and suggests training pipeline design as a key differentiating factor. Verbalized and token-aggregation sequence-level methods fail systematically. Only semantic consistency, as measured by consistency of the final answer, yields well-calibrated outputs, providing the first large-scale evidence that semantic calibration persists in multi-step, dependency-rich reasoning settings.
\end{abstract}

\begin{figure*}[t]
    \centering
    \includegraphics[width=\linewidth]{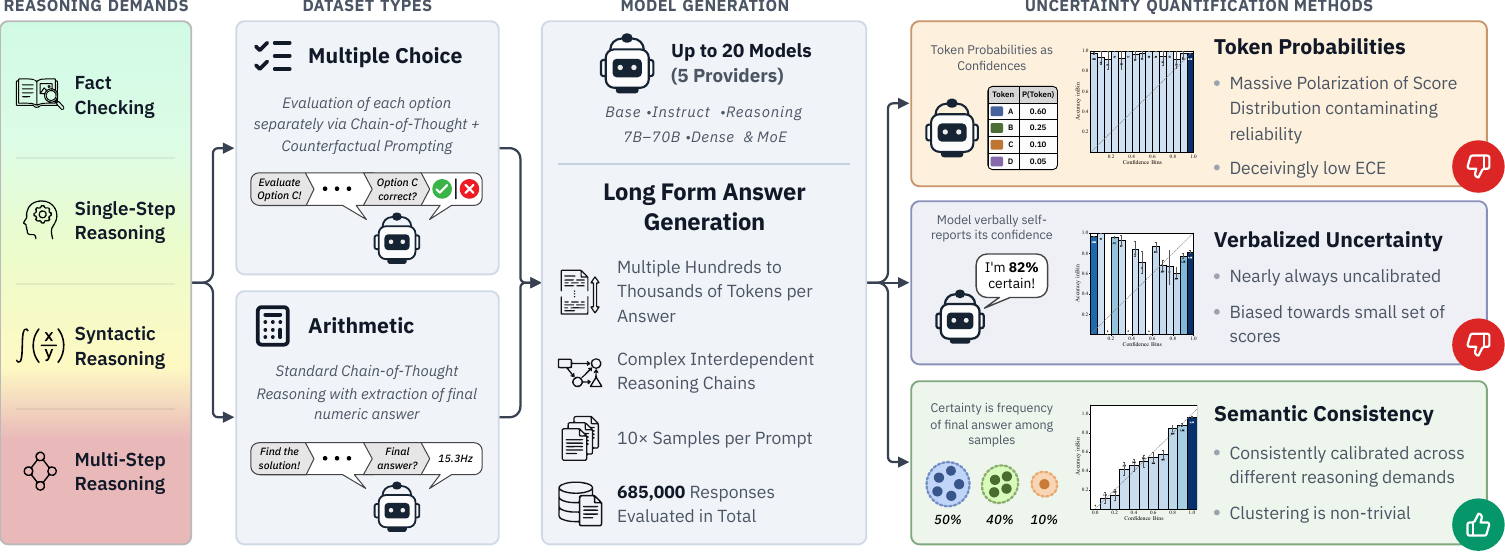}
    \caption{\textbf{Overview of the Benchmark and Evaluation Pipeline.} Diverse reasoning demands and QA task formats are used to generate large-scale long-form responses from up to 20 language models. Four uncertainty quantification methods are evaluated under increasing reasoning complexity, revealing consistent calibration only for semantic consistency while token probabilities and verbalized uncertainty remain poorly calibrated.}
    \label{fig:main_figure}
\end{figure*}

\section{Introduction}
\label{sec:intro}

Large language models (LLMs) are increasingly deployed for complex question answering tasks spanning knowledge-intensive scientific queries, multi-step reasoning problems, and research question answering. These LLMs now form core components of search-augmented agentic reasoning pipelines \citep{jin2025searchr1trainingllmsreason}, automated knowledge-discovery systems \citep{xiong2024search}, and AI-powered search \citep{kelly2023bing}. In these settings, long-form outputs with interdependent reasoning steps are generated at a scale that exceeds practical human verification. Automated UQ is therefore essential for flagging unreliable outputs and mitigating error propagation in downstream pipelines. Undetected propagated uncertainties can lead to severe downstream errors that may remain unnoticed in scientific QA, including in agentic research assistance \citep{Gu_2025}, automated research direction discovery \citep{lu2024aiscientist,kramerAutomatedScientificDiscovery2026}, and long-form information synthesis.

Calibration characterizes whether a model’s stated confidence reflects its actual likelihood of being correct (accuracy), allowing confidence scores to support reliable automated decisions and risk-aware system behavior \citep{guo2017calibration}. This task is orthogonal to selective prediction \citep{xia2026whatandwhat}, which focuses on identifying individually unreliable generations for abstention under risk–coverage trade-offs. Selective prediction has received substantially more benchmarking attention \citep{fadeeva-etal-2023-lm}.

Token-level confidence scores are the most commonly deployed UQ signal, but their reliability in instruction-tuned and reasoning models remains poorly understood at scale \citep{2303.08774}. When these scores systematically collapse through instruction tuning, any downstream process relying on confidence-based filtering receives near-constant high-confidence signals regardless of actual answer quality, a failure mode invisible without distribution-level analysis.

\paragraph{Scientific QA as a benchmark domain.}
Scientific QA spans a full reasoning spectrum, from fact retrieval to multi-step symbolic reasoning, producing long-form answers with interdependent reasoning steps. Answers are often too complex for casual verification, making UQ most needed and least validated precisely here. Prior calibration benchmarks target short-form or classification-style outputs \citep{guo2017calibration,2303.08774,fadeeva-etal-2023-lm}. This benchmark instead evaluates calibration on unrestricted multi-sentence reasoning chains. Verifiable ground truth in the form of multiple choice question answering (MCQA) labels or arithmetic answers enable automatic computation of calibration metrics without costly human annotation.

\paragraph{Research questions.}
This paper addresses two research questions:
\begin{enumerate}[label=\textbf{(RQ\arabic*)}, leftmargin=*, topsep=4pt]
    \item To what extent are token-level probabilities a calibrated measure of model confidence, and what effect does instruction tuning or the reasoning process have on calibration?
    \item To what extent do sequence-level UQ methods provide reliable uncertainty estimates for long-form answers in scientific QA, across tasks spanning retrieval-based to reasoning-intensive questions?
\end{enumerate}

\paragraph{Contributions.}
\textbf{First}, we introduce the first large-scale calibration benchmark for long-form scientific QA, comprising \samplecount{} responses across up to 20 LLMs (base, instruction-tuned, and reasoning variants) and seven datasets, with a reproducible shared-generation design in \texttt{async-graph-bench}, an open-source framework developed for this paper. \textbf{Second}, we characterize instruction-tuning-induced probability mass polarization across five model families and four MCQA datasets; reasoning families diverge by provider in their response to this effect (see \Cref{sec:token_calib}; all findings are observational). \textbf{Third}, we show that verbalized and token-aggregation methods fail systematically while semantic consistency via answer consistency yields well-calibrated outputs, extending semantic calibration into multi-step, dependency-rich reasoning settings.

\section{Related Work}
\label{sec:rel_work}

\paragraph{Token-level UQ and calibration.}
UQ for LLMs addresses the hallucination problem intrinsic to token-level probabilistic generation \citep{2005.00661,2406.04175,2309.01219,2409.05746v1}. A broad range of UQ methods has been established for LLM-generated text \citep{geng_survey_2024}, though many assume equal token importance and thereby misrepresent sequence-level confidence \citep{2502.04955,2302.09664}. Empirical evidence consistently shows that base LLMs are better calibrated than their instruction-tuned counterparts, as measured by Expected Calibration Error (ECE) \citep{2303.08774,2305.14975,wang2025objectivefinetuningllmsprior}. The GPT-4 Technical Report \citep{2303.08774} documents ECE degradation after instruction tuning without analyzing the underlying score distribution shift. \citet{2407.14614} observed polarization of label probability scores due to instruction tuning in a narrow binary classification setting, leaving generalizability across model families and diverse reasoning tasks open.

\paragraph{Benchmarking calibration in LLMs.}
LM-Polygraph \citep{fadeeva-etal-2023-lm} is the primary UQ benchmarking framework for LLMs, evaluating selective prediction via the Prediction Rejection Ratio (PRR), but does not evaluate calibration and targets short-form outputs. \citet{tao2025revisitinguncertaintyestimationcalibration} conduct a broad model comparison restricted to a single dataset; \citet{2503.15850} survey UQ methods without empirical calibration comparison. Calibration-focused evaluation across diverse model families, reasoning-demanding tasks, and long-form generation remains poorly characterized at scale.

\paragraph{Semantic calibration.}
\citet{semanticcalibration} demonstrate that LLMs exhibit strong calibration at the level of semantic answer clusters despite token-level miscalibration, but restrict their analysis to short, single-hop answers. Whether semantic calibration persists in multi-step, dependency-rich reasoning settings remains open, and is directly addressed by our benchmark. No prior work simultaneously evaluates calibration across long-form scientific QA, multiple UQ paradigms, and large-scale model diversity spanning base, instruction-tuned, and reasoning models across five providers.

\section{Methodology}
\label{sec:methodology}

\subsection{Benchmark Design}
\label{ssec:design_criteria}

A fundamental challenge in cross-method UQ comparison is that integrating new methods either requires rerunning the entire benchmark across all methods, which is computationally costly, or evaluating the new method on separately generated outputs. In the latter case, stochastic decoding prevents performance differences from being cleanly attributed to the UQ method itself rather than to generation variance, thereby compromising comparability. We therefore evaluate all UQ methods on identical shared generations: outputs are generated once, cached, and reused across all evaluations. We developed \texttt{async-graph-bench}, an open-source framework for reproducible multi-stage UQ benchmarking, released alongside the computed UQ scores and full reproduction instructions at \repositorylink{}.

Scientific QA is chosen as the target domain because its dependency-rich reasoning chains expose compounding uncertainty: errors propagate across dependent steps in ways short-form benchmarks conceal, making it an ideal stress test for UQ methods. Crucially, MCQA structure is used exclusively for automatic answer verification, not to constrain output format; APriCoT prompting \citep{2408.08651v2} elicits open-ended multi-sentence reasoning traces and arithmetic tasks impose no length restrictions.

\subsection{Dataset Selection}
\label{ssec:datasets}

The requirement for verifiable ground truth to compute calibration metrics (ECE, AUROC) drives use of structured tasks: four MCQA datasets and three arithmetic QA datasets, with emphasis on scientific QA covering physics, chemistry, and biology, plus mathematics for a low-complexity arithmetic baseline. Datasets span three reasoning tiers (full details in the appendix):

At the \textbf{recall tier}, MMLU \citep{2009.03300}, SciQ \citep{1707.06209}, and ARC-Easy \citep{1803.05457} test factual recall and single-step reasoning. At the \textbf{multi-step tier}, ARC-Challenge \citep{1803.05457}, GSM8K \citep{2110.14168} with its multiple-choice variant GSM-MC \citep{2405.11966}, and SVAMP \citep{2103.07191} require chained reasoning steps. At the \textbf{high-complexity tier}, SciBench \citep{2307.10635} requires formula manipulation, unit tracking, and multi-equation operations; GPQA \citep{2311.12022} demands expert multi-hop reasoning explicitly designed to resist lookup.

Response lengths, as reported in \autoref{tab:token_lengths_combined}, confirm the long-form nature of the generated outputs. The average mean response length per dataset ranges from 286 tokens on SVAMP to 2,403 tokens on GPQA. On the most challenging datasets, this increases substantially for reasoning models, with mean response lengths averaged across models reaching 4,194 tokens on GPQA and 4,086 tokens on SciBench. Even instruction-tuned models exceed 1,000 tokens on average for these datasets (GPQA: 1,123 tokens; SciBench: 1,029 tokens).

\subsection{Model Selection}
\label{ssec:models}

Models are chosen to cover a broad and up-to-date LLM spectrum while ensuring reproducibility via public open-weight models. The selection spans five providers (OpenAI, Mistral, Meta, Qwen, and Google), varying sizes (7B to 70B parameters), three variant types (base, instruction-tuned, and reasoning), and multiple architectures including mixture-of-experts variants. Instruction-tuned and reasoning models are complemented by their base model counterparts to enable controlled comparison of instruction-tuning and reasoning-training effects.

\subsection{First Experiment: Token-Level Label Probability Calibration}
\label{ssec:exp1_setup}

The first experiment replicates and substantially extends the token-level calibration analysis from the GPT-4 Technical Report \citep{2303.08774}, expanding along three dimensions. First, model coverage is extended to base, instruction-tuned, and reasoning model variants across all five providers, rather than a single model. Second, dataset diversity spans four MCQA datasets covering the full reasoning complexity spectrum: MMLU (factual recall), ARC-Challenge (multi-step reasoning), GSM8K-MC (symbolic arithmetic), and GPQA (expert multi-step reasoning). Third, analysis depth includes calibration plots and score distribution analysis alongside ECE, characterizing what causes calibration degradation rather than only observing that it occurs.

Models are prompted to respond with a single answer label (A, B, C, or D) for each MCQA item. Token-level probabilities assigned to these label tokens serve as confidence scores. Reasoning models produce extended reasoning traces, but the final label token probability is the confidence measure. Normalization of token probabilities over answer labels is applied and validated following \citet{2402.01349}.
\begin{figure*}[t]
    \centering
    \includegraphics[width=\linewidth]{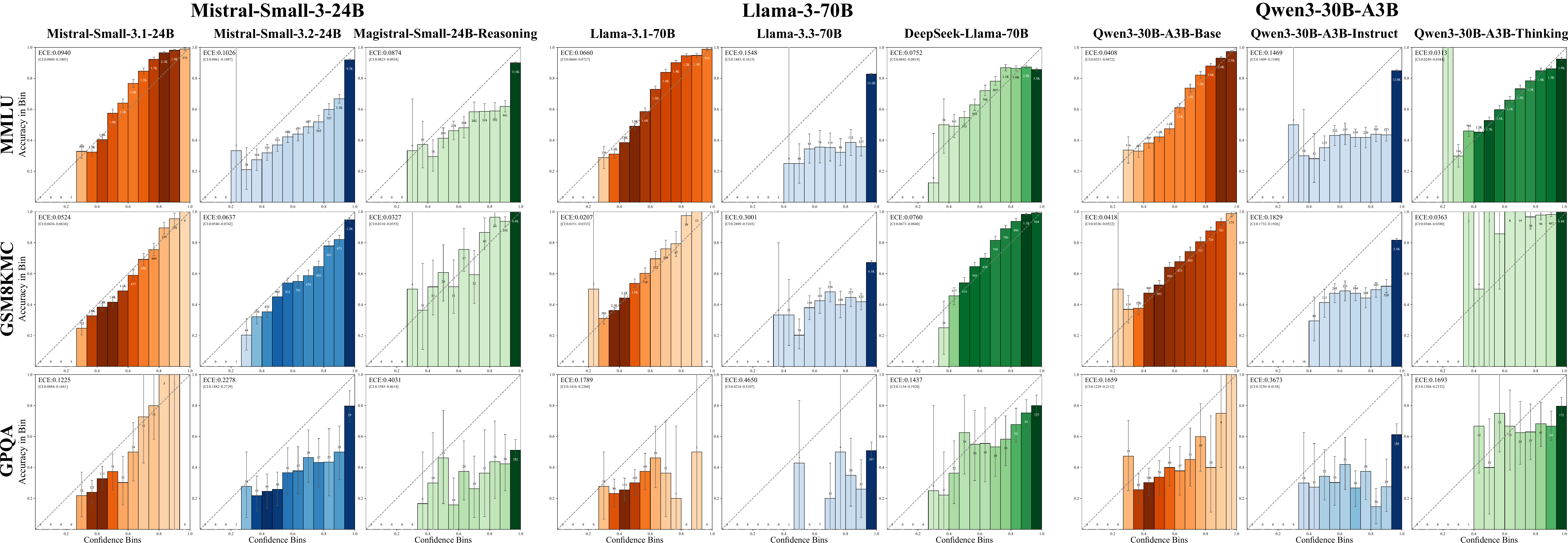}
    \caption{\textbf{Calibration Plots Using Label Probabilities as Uncertainty Scores.} Columns correspond to three selected model families: base variants in orange, instruction-tuned in blue, reasoning in green. Rows refer to MMLU, GSM8KMC, and GPQA. Darker shading indicates more items per confidence bin; each bin is labelled with its sample count. Results use Prompt Design 1 with structured decoding enabled.}
    \label{fig:mcqa_calibration_prompt1}
\end{figure*}
\subsection{Second Experiment: Sequence-Level UQ Calibration}
\label{ssec:uq_methods}

Our second experiment evaluates sequence-level UQ calibration across all seven datasets. It is restricted to instruction-tuned and reasoning models, as base models exhibit insufficient task comprehension for long-form QA. All four UQ methods are applied to identical cached long-form generations. Metric implementation details can be found in \autoref{sec:metric_implementation}.

\paragraph{Method selection.}
Calibration requires normalized per-instance confidence scores in $[0,1]$; methods producing rankings, unnormalized scores, or claim-level outputs are excluded. The four selected methods cover three paradigms: token-probability aggregation, self-evaluation, and semantic consistency. A comprehensive method overview, relevance statement and selection rationale are provided in \Cref{app_ssec:method_citation,app_ssec:unc_method_exclusion}.

\paragraph{Verbalized Uncertainty.}
\citet{2305.14975} propose prompting the model post-generation for a self-assessed probability of correctness. This method is straightforward but ignores token probability distributions and may be susceptible to training data bias.

\paragraph{P(True).}
\citet{2207.05221} propose prompting the model post-generation to classify its own answer as \token{(A)} True or \token{(B)} False. Token probabilities of these two labels are used as confidence scores.

\paragraph{Claim-Conditioned Probability (CCP).}
\citet{2403.04696} propose CCP, which estimates token-level uncertainty by clustering the top probable token alternatives into entailing and contradicting classes using natural language inference (NLI). Sequence confidence is computed as the product of per-token confidences.

\paragraph{Answer Consistency.}
\citet{2203.11171} propose measuring the proportion of semantically equivalent answers across multiple sampled generations. For MCQA datasets, a follow-up prompt assesses whether the generation concludes that the evaluated answer option is correct; responses are clustered per answer option by this binary label. For arithmetic QA, a follow-up prompt extracts the final numeric answer from each generation; responses are clustered by the extracted value. This design separates the initial long-form generation from the answer-mapping step used for ground-truth comparison. All confidence scores are computed from agreement among the sampled judgments per answer option alone; gold labels are introduced only afterwards to determine the correctness of each judgment and to evaluate calibration (\Cref{app_ssec:answer_consistency_definitions,sec:metric_implementation}). This method is computationally expensive but proxies self-consistency prompting.

\subsection{Prompting Strategy}
\label{ssec:prompting}

MCQA introduces selection bias and positional bias, including systematic preferences for certain label positions \citep{2309.03882} and token and formatting frequency bias \citep{2406.07545}. APriCoT prompting \citep{moore2025chainthoughtthinksfast} addresses both biases by evaluating each answer choice independently as correct or incorrect via Chain-of-Thought combined with counterfactual prompting. This preserves a verifiable label format while eliciting genuine long-form reasoning responses. APriCoT produces four separate prompts per MCQA item, one per answer option. The MCQA structure constrains ground truth verification only, not response format; responses are multi-sentence reasoning chains, not single label tokens. For Arithmetic QA we use Chain-of-Thought (CoT) prompting.

Each dataset is subsampled to 250 items with 10 generations per prompt, yielding $2{,}500$ responses per arithmetic dataset and $10{,}000$ per MCQA dataset via APriCoT ($250 \times 4 \times 10$); full counts (\samplecount{} total) are detailed in \Cref{appendix:prompt_composition}.

For the first experiment, four alternative 3-shot prompts are designed to address task comprehension differences between base and instruction-tuned models. The prompt maximizing average probability mass on label tokens is selected per model, as a proxy for task comprehension. Structured decoding is used to reliably obtain relative label probabilities and prevent invalid answers.

\subsection{Evaluation Metrics}
\label{ssec:eval_metrics}

Expected Calibration Error (ECE) is the most widely used aggregate metric for calibration evaluation \citep{guo2017calibration,hendrycks2018baseline}. However, as a coarse summary statistic, it can obscure important distributional pathologies. In particular, when confidence estimates collapse into a narrow or highly polarized range, ECE becomes dominated by overall model accuracy rather than reflecting whether confidence scores remain informative at the instance level. This limitation is consistent with the broader observation that summary statistics can conceal substantial structural differences in the underlying data distribution \citep{matejka2017same}. Consequently, aggregate calibration metrics alone are insufficient for a comprehensive assessment of calibration behavior.

\paragraph{ECE failure under polarization.}
When probability mass collapses, as documented in the binary classification setting by \citet{2407.14614} and extended here across model families, confidence scores become effectively degenerate. If the collapsed score range coincides with the model's average accuracy, ECE may remain deceptively low despite the confidence estimates no longer supporting meaningful thresholding, selective prediction, or risk-aware decision making. We therefore pair calibration plots with ECE throughout this benchmark to preserve visibility into the underlying score distributions and calibration structure.

To assess robustness, we estimate bootstrap confidence intervals at the 95\% confidence level over calibration curves and representative ECE values using 1000 resamples. AUROC values are reported in the full result plots and tables for completeness and to make our benchmark results amenable to the selective prediction community. As the primary focus of this work is calibration, AUROC is not used as an evaluation criterion.

\section{Results: Token-Level Calibration}
\label{sec:token_calib}

\begin{figure*}[t]
    \centering
    \includegraphics[width=\linewidth]{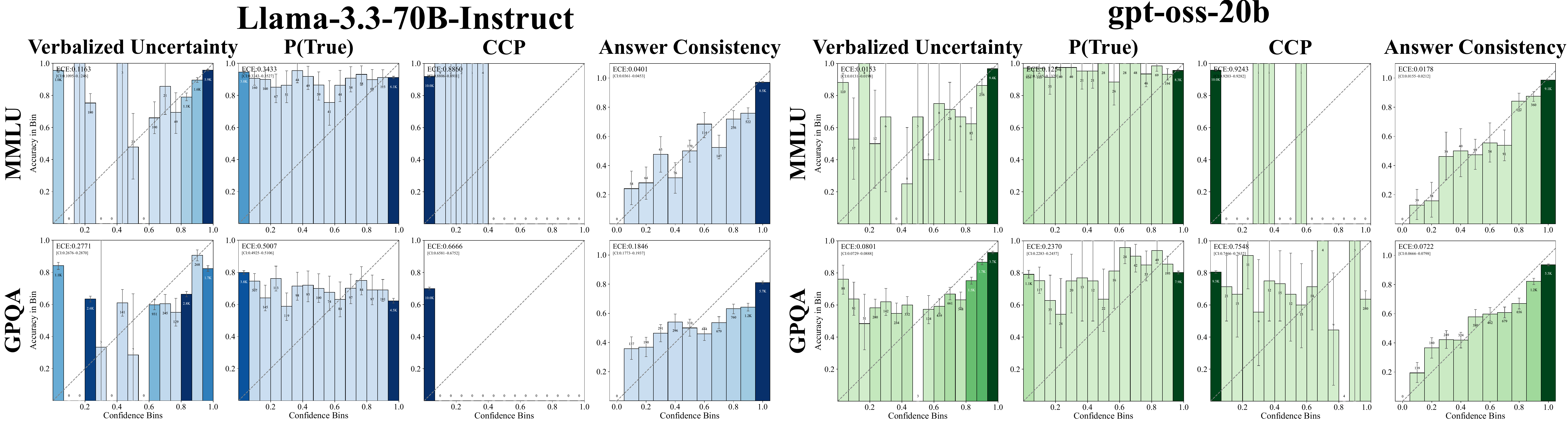}
    \caption{\textbf{Selected Calibration Plots for the Four Evaluated Sequence-Level UQ Methods.} Results for \model{Llama-3.3-70B} (left) and \model{gpt-oss-20b} (right) on MMLU and GPQA.}
    \label{fig:calib_exp2}
\end{figure*}

\subsection{Main Findings}
\label{ssec:token_main}
\autoref{fig:mcqa_calibration_prompt1} reports a subset of representative calibration plots. All results shown use Prompt Design 1 (see \Cref{sec:exp1_prompt_designs}), which was found to optimize task comprehension across model types (see \Cref{sec:task_comprehension_prompts}). Structured decoding has been employed to generate the final result, after normalization has been validated (see \Cref{app_sssec:normeff}).
Comprehensive results for all models, datasets and prompt designs are included in \Cref{sec:exp1_full_plots}.

ECE increases with reasoning complexity: symbolic and multi-step tasks such as GSM8K and GPQA exhibit substantially higher ECE than factual retrieval tasks such as MMLU. Token-level probabilities capture aleatoric uncertainty in fact retrieval but become unreliable for reasoning tasks requiring epistemic uncertainty estimation across multiple dependent steps.

Instruction tuning degrades ECE across most model families. Calibration plots reveal systematic polarization: many instruction-tuned models concentrate nearly all probability mass on a single label, rendering confidence scores near-constant across items and destroying their value as per-instance reliability signals. This polarization is characterized across five model families and four MCQA datasets spanning the full reasoning complexity spectrum; Mistral models are a notable exception.

\subsection{Reasoning Model Divergence}
\label{ssec:reasoning_divergence}

Reasoning models diverge markedly across providers. \model{gpt-oss} and \model{Magistral} commit to a single answer option early in the reasoning process, and token probabilities collapse as in instruction-tuned models. The reasoning trace does not distribute probability mass across alternatives, resolving uncertainty prematurely rather than exploring multiple possibilities.

\model{DeepSeek-R1} and \model{Qwen3}, by contrast, actively explore and assess multiple answer options in their reasoning traces, externalizing uncertainty via epistemic markers across alternatives. Token probabilities remain distributed, and calibration is substantially preserved or improved relative to base models, at the cost of considerably longer reasoning traces.

These differences cluster by provider rather than by scale or architecture. Training pipeline design, including RLHF targets, CoT supervision, and reward shaping, is the most plausible differentiating factor. Without access to training configurations, however, this remains an observational hypothesis rather than a causal claim. The benchmark provides a foundation for further investigation into how individual training stages influence model calibration and polarization.

\section{Results: Sequence-Level Calibration}
\label{sec:seq_calib}

\begin{figure}[t]
    \centering
    \includegraphics[width=\linewidth]{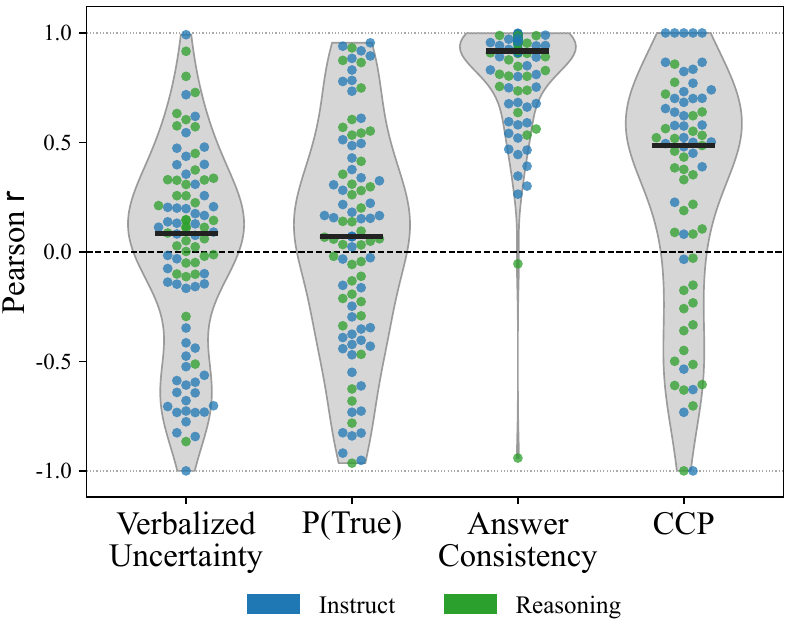}
\caption[Calibration Correlation by UQ Method]{\textbf{Calibration Correlation by UQ Method.} Distribution of Pearson calibration correlation ($r$) across all (model, dataset) pairs for each UQ method. Correlations are computed between the bin-confidence midpoints and the empirical bucket accuracies of the corresponding calibration curves, excluding empty bins and weighting all non-empty bins equally. Higher values indicate stronger alignment between confidence and empirical accuracy. Violins summarize the overall distribution and horizontal bars indicate the median.}
    \label{fig:calibration_correlation}
\end{figure}

\autoref{fig:calib_exp2} reports a representative subset of calibration plots. Extensive visualizations for all models and datasets across UQ methods are provided in \Cref{sec:extensive-calibration-plots-exp2}. \autoref{fig:calibration_correlation} reports the distribution of calibration correlation across all (model, dataset) pairs for each UQ method. Additional aggregate comparisons based on ECE and AUROC are reported in \Cref{fig:calibration_ece_swarm,fig:calibration_auroc_swarm}. The following subsections provide method-wise analyses of these calibration behaviors.

\subsection{Verbalized Uncertainty}

\begin{figure}[t]
    \centering
    \begin{subfigure}[b]{0.48\textwidth}
        \centering
        \includegraphics[width=\textwidth]{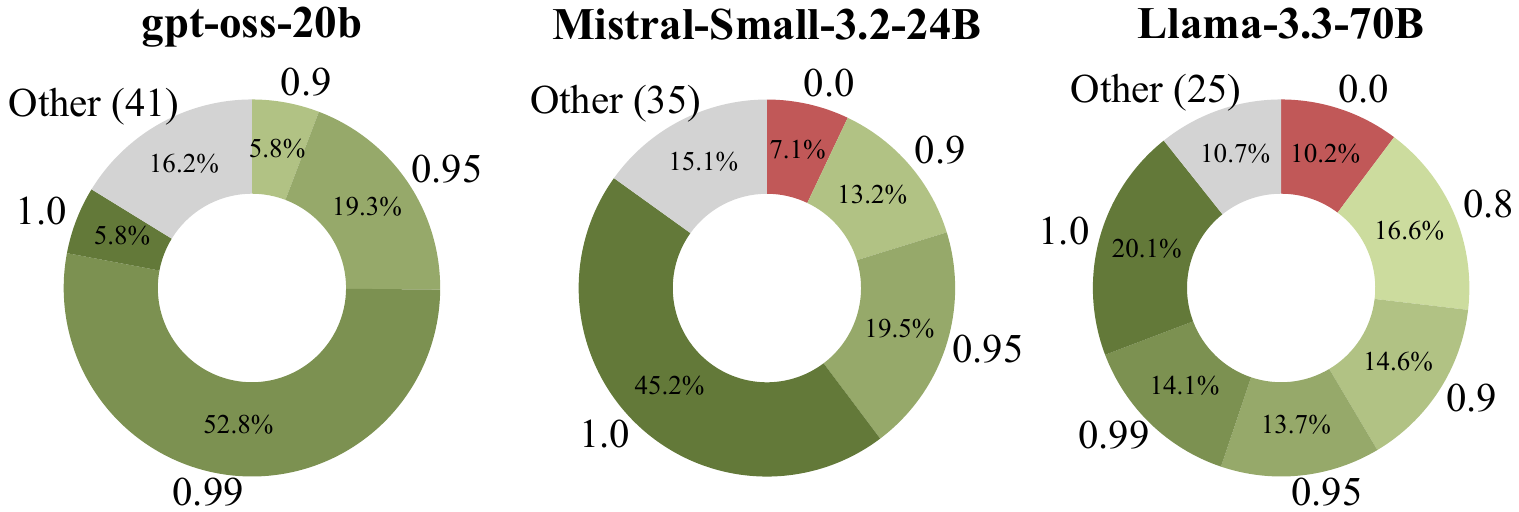}
        \caption{\textbf{Verbalized Uncertainty.} Confidence scores seen in less than $5\%$ of total responses are grouped into ``Other'', with the number of distinct scores in brackets.}
        \label{fig:verbalized}
    \end{subfigure}
    \hfill
    \vspace{0.01cm}
    \begin{subfigure}[b]{0.48\textwidth}
        \centering
        \includegraphics[width=\textwidth]{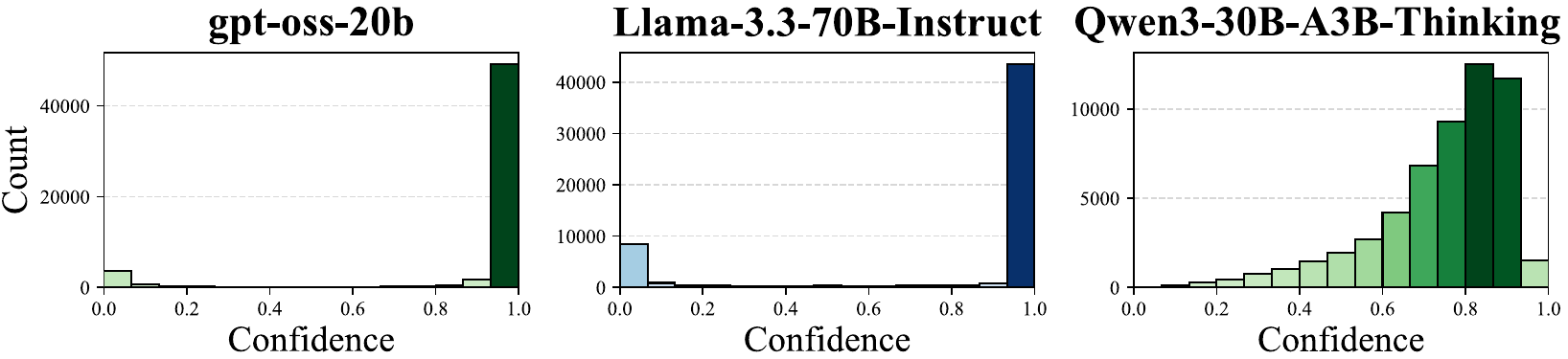}
        \caption{\textbf{P(True).} Confidence scores are derived from probability mass assigned to labels \token{(A)} and \token{(B)}, representing the model's confidence that the answer is true or false respectively.}
        \label{fig:ptrue}
    \end{subfigure}
    \caption{\textbf{Distribution of Confidence Scores Across Selected Representative Models.} Scores aggregated across all datasets ($57{,}500$ prompts total). See \Cref{sec:verbalized-uncertainty-distribution} and \Cref{sec:ptrue-distribution} for extensive plots across all models.}
    \label{fig:both}
\end{figure}

Models default to a small set of confidence scores, with high-confidence scores dominating the output distribution, as seen in \Cref{fig:verbalized}. This biased distribution is associated with training data composition and instruction tuning. Calibration plots show no correlation between verbalized scores and answer accuracy across instruction-tuned models.

A slight improvement is observed in reasoning models, most notably the \model{gpt-oss} family, which achieves better correlation with correctness on challenging datasets such as GPQA and SciBench. Without access to training configurations and workflow setups, we can only speculate that this reflects training-related factors specific to this model family.

The failure mode is characterized by output distribution collapse driven by training data bias: instruction-tuned models default to high-confidence scores regardless of actual answer correctness. This is not a calibration mismatch in the statistical sense but a fundamental collapse of the output distribution, making verbalized uncertainty unreliable as a per-instance uncertainty signal.

\subsection{P(True)}

Many models assign near-1.0 confidence scores with little use of intermediate values, resulting in a polarized distribution, as seen in \Cref{fig:ptrue}. Calibration plots show no meaningful correlation between P(True) scores and actual correctness across models.

P(True) relies on token-level probabilities for its confidence signal. The polarization documented in \Cref{sec:token_calib} directly propagates: when token probabilities collapse, P(True) assigns near-constant high confidence regardless of answer correctness. The failure of P(True) is thus a downstream consequence of token-level polarization, providing a direct cross-experiment link between the findings of \Cref{sec:token_calib} and those of this section.

\subsection{Claim-Conditioned Probability (CCP)}

CCP suffers from vanishing sequence-level confidence scores: multiplying many per-token confidences drives scores toward zero regardless of correctness, yielding uninformative near-zero signals for all long-form outputs. Calibration plots show no meaningful alignment with correctness and high calibration error across models or datasets. Three structural failure modes are identified: (1) vanishing scores due to cumulative probability multiplication; (2) NLI misclassification instability, where a single incorrect entailment decision propagates multiplicatively and dominates the sequence score; and (3) stop-word contamination, where grammatical tokens dilute the semantic signal during aggregation. Each represents a concrete direction for future improvement.

\subsection{Answer Consistency}

Higher answer frequencies strongly correlate with correctness across both MCQA and arithmetic tasks. More challenging datasets such as GPQA and SciBench exhibit greater answer diversity across sampled generations, reflecting higher model uncertainty on these tasks.

Calibration plots confirm well-aligned confidence estimates across model families and dataset types, as seen in \Cref{sec:calibration-correlation}, making Answer Consistency the most reliable approach evaluated in this benchmark. Semantic calibration persists in long-form, multi-step, dependency-rich reasoning settings where all other methods collapse. This extends and validates the finding of \citet{semanticcalibration} into a substantially more challenging reasoning regime: answer consistency remains the most reliable and well-calibrated uncertainty indicator even in complex scientific QA requiring extended reasoning chains.

\paragraph{How Answer Consistency differs from single-trajectory methods.}
Answer Consistency differs in kind from the other evaluated methods: while token probabilities and verbalized confidence read uncertainty off a single generation trajectory, it measures agreement across independently generated semantic outcomes, i.e., whether the model consistently converges to the same semantic answer rather than how confident it appears along one generation path. We interpret this contrast as a plausible reason for the observed pattern, in which single-trajectory signals collapse after instruction tuning while Answer Consistency remains calibrated; the benchmark is observational, and we do not claim to have established this mechanism.

Answer Consistency is not universally reliable. Three conditions can undermine it: \emph{systematic shared errors}, in which a model repeatedly produces the same incorrect answer and thus assigns high confidence to an incorrect response; \emph{ambiguous questions} that admit multiple valid interpretations, in which case diverse answers reflect genuine task uncertainty rather than model miscalibration; and \emph{errors in semantic equivalence clustering}, in which the answer-mapping step merges distinct answers or splits equivalent ones. Our results indicate that none of these conditions is systematically in force on the structured tasks evaluated here, which is consistent with the well-calibrated behavior observed across models and datasets; the clustering condition nonetheless becomes harder to guarantee for open-ended QA. See \Cref{app_ssec:answer_consistency_definitions} for the formal definitions and the Limitations section for a discussion of scope.

The primary limitation is computational cost, which scales linearly with the number of samples per prompt (10 samples per prompt in this benchmark; \Cref{appendix:prompt_composition}). Answer Consistency is restricted to settings where cluster formation is tractable. For practitioners requiring calibrated confidence estimates in scientific or reasoning-intensive QA, Answer Consistency is the most reliable option among currently evaluated methods; token-based methods (CCP, P(True)) and verbalized uncertainty should not be used as calibration quality signals without further validation.

\section{Discussion}
\label{sec:discussion}

\subsection{Implications of Token-Level Polarization}
\label{ssec:polarization_implications}

Polarization is not identical to overconfidence. Overconfidence denotes a systematic tendency for confidence scores to exceed actual accuracy; polarization denotes a full collapse of the score distribution. A polarized model has lost any discriminative signal, and post-hoc recalibration cannot recover information no longer present in the scores: standard procedures such as temperature scaling or Platt scaling presuppose that scores retain some ordinal relationship with correctness. Practitioners relying on token-level confidence thresholds for abstention or routing in reasoning-intensive domains should therefore verify that scores retain discriminative value before deployment.

\subsection{Reasoning Models and Calibration}
\label{ssec:reasoning_discussion}

The provider-clustered divergence documented in \Cref{ssec:reasoning_divergence} suggests that training pipeline design, not the mere presence of a reasoning process, is the key determinant of calibration quality. The behavioral contrast between polarization-mitigating models (\model{DeepSeek-R1}, \model{Qwen3}) and polarization-reproducing ones (\model{gpt-oss}, \model{Magistral}) is consistent with differences in reasoning supervision: exploratory multi-option deliberation during the reasoning trace appears to preserve probability mass distribution. This interpretation is observational; controlled ablation of specific training stages is a direct open problem uncovered by this benchmark.

\section{Conclusion}
\label{sec:conclusion}

Our large-scale evaluation of four UQ methods across 20 models demonstrates that semantic calibration persists in complex reasoning tasks, yet token-level probabilities are largely uninformative due to polarization induced by instruction tuning. Although reasoning processes can attenuate this effect, the strongest differences are observed across providers rather than architectures. Verbalized uncertainty and CCP suffer from distribution collapse and instability, whereas Answer Consistency proves reliable despite higher computational cost. Our findings underscore the necessity of evaluating UQ at the semantic outcome level~\citep{semanticcalibration} and call for sustained benchmarking to disentangle the contributions of architecture, fine-tuning, and training data to calibration behavior.

Future work should extend the evaluation to additional UQ methods and study the effects of inference parameters and prompting strategies on calibration. Reformulating MC tasks as open-ended would isolate task-specific uncertainty sources. Controlled ablation of individual training stages (SFT, RLHF, RLAIF, and inference-time scaling) would determine whether the provider-clustered behavioral differences observed here are driven by specific pipeline components, transforming an observational finding into an actionable mechanism for improving model calibration.

\section*{Limitations}
\label{sec:limits}

Despite approximating open question answering for MCQA via APriCoT prompting and complementing with arithmetic QA, the benchmark scope remains restricted to structured task formats. This is a deliberate design choice: calibration metrics such as ECE require unambiguous correctness labels, and introducing LLM-judged or human-evaluated open-ended correctness would add a second source of uncertainty to the measurement pipeline, making it impossible to disentangle miscalibration in the model's self-assessment from noise in the evaluation method itself. Even so, the structured-task scope limits generalizability to some extent, e.g. to fully open-ended generation and other task types. All datasets are in English, leaving open questions about behavior in multilingual or code-generation settings.

Despite APriCoT mitigation, the MC setting may introduce steering effects on model responses. Arithmetic QA provides complementary evidence with fully open-ended numerical answers, and the consistency of findings across both task types and model variants supports generalizability. Explicit cross-domain validation nonetheless remains necessary.

Answer Consistency validity is contingent on the quality of semantic equivalence clustering. This is tractable for MCQA and arithmetic QA but non-trivial for open-ended QA.

Our benchmark cannot establish why instruction tuning induces polarization or why certain reasoning models mitigate it. We speculate that provider-level behavioral clustering is due to training pipeline differences. Controlled ablations were not feasible as training pipeline setups are undisclosed by model providers.

Inference parameters are held constant, including fixed temperature and fixed decoding settings across all models and datasets. Variation in these parameters is not examined in this study.

\bibliographystyle{acl_natbib}
\bibliography{master_bibliography}

\renewcommand{\thetable}{A.\arabic{table}}
\renewcommand{\thefigure}{A.\arabic{figure}}
\renewcommand{\theequation}{A.\arabic{equation}}
\setcounter{table}{0}
\setcounter{figure}{0}
\setcounter{equation}{0}

\newpage
\appendix

\section{Appendix}
\raggedbottom
\subsection{Reproducibility Statement}
Reproducibility in UQ for LLMs poses unique challenges due to the multistage evaluation pipelines and inherent randomness at each step, which can compound and introduce measurement noise. This stochasticity affects comparisons across UQ methods and model outputs, particularly when methods are evaluated on different generations, potentially contaminating results. To address these issues, we developed a modular benchmarking framework designed to ensure reproducibility and resource efficiency in large-scale LLM experiments. The framework supports extensible and replaceable computation nodes, caches probabilistic outputs such as model generations to ensure consistent re-evaluation, preserves intermediate outputs for qualitative inspection and allows incremental updates so that only affected steps need recomputation. It also enables sharing of intermediate results to reduce compute costs and allow other researchers to rerun the benchmark with their own methods. This is particularly valuable for ongoing research in UQ on models gated by proprietary access or costly hardware, supporting broader collaboration within the scientific community. These features not only enhance the reliability of the present study but also establish a foundation for future research in the field.

To facilitate reproducibility, the repository (see \Cref{app:sec_repository}) includes: the complete \texttt{async-graph-bench} framework with extensible, replaceable computation nodes; final leaf-node outputs containing all confidence scores required to reproduce every plot and table in this paper; the implementation files of all four UQ methods, with hyperparameters visible in code (e.g., \lstinline{models.py}, \lstinline{apricot\_mc\_calc.py}, \lstinline{ptrue.py}, \lstinline{arithmetic\_calc.py}); a container definition file to build the execution environment; exact pip and driver versions pinned for deterministic reproduction; and installation and benchmark-execution instructions. All four UQ methods are evaluated on the identical cached generations, eliminating generation variance from cross-method comparisons. The main barrier to full reproduction is therefore the computational cost of generating the \samplecount{} responses, for which we release the resulting confidence scores instead.


\subsection{Repository}
\label{app:sec_repository}
The repository accompanying this paper is available online at \repositorylink{}. It contains the framework instance used for this benchmark, the computed UQ scores (final leaf-node outputs), method implementations, the container definition, and pinned dependency versions, as detailed in the Reproducibility Statement. The underlying benchmarking framework, \texttt{async-graph-bench}, is developed and maintained separately at \frameworklink{}.

\subsection{Information About the Use of AI Assistants}
AI assistants were used to support text refinement, including improvements to wording, phrasing, and overall readability of the manuscript. AI-based tools were also used for coding assistance during implementation and experimentation. All generated content, code, and suggestions were reviewed and verified by the authors before inclusion in this work.

\subsection{Model Selection}
\label{app_ssec:model_sel}

An overview of the models used in the experiments is given in \Cref{tab:updated-llm-selection}; primary technical references per model family are listed in \Cref{tab:model_families}.
\begin{table*}[t]
  \centering
  \renewcommand{\arraystretch}{1.1}
  \begin{tabular}{l l c r c}
    \hline
    \textbf{Provider} & \textbf{Model Name} & \textbf{Type} & \textbf{Size} & \textbf{Release Date} \\
    \hline
    \multirow{2}{*}{OpenAI} 
      & gpt-oss-20b & Reasoning & 20B & Aug 2025 \\
      \cline{2-5}
      & gpt-oss-120b & Reasoning & 120B & Aug 2025 \\
    \hline
    \multirow{7}{*}{Mistral AI} 
      & Mistral-Nemo-Base-2407 & Base & 8B & Jul 2024 \\
      & Mistral-Nemo-Instruct-2407 & Instruct & 8B & Jul 2024 \\
    \cline{2-5}
      & Ministral-8B-Instruct-2410 & Instruct & 8B & Oct 2024 \\
      \cline{2-5}
      & Mistral-Small-3.1-24B-Base-2503 & Base & 24B & Mar 2025 \\
      & Mistral-Small-3.2-24B-Instruct-2506 & Instruct & 24B & Jun 2025 \\
      & Magistral-Small-2507 & Reasoning & 24B & Jul 2025 \\
    \hline
    \multirow{4}{*}{Meta LLaMA} 
      & Llama-3.1-70B & Base & 70B & Jul 2024 \\
      & Llama-3.3-70B-Instruct & Instruct & 70B & Dec 2024 \\
      \cline{2-5}
      & Llama-4-Scout-17B-16E & Base & 109B & Apr 2025 \\
      & Llama-4-Scout-17B-16E-Instruct & Instruct & 109B & Apr 2025 \\
    \hline
    \multirow{3}{*}{Qwen} 
      & Qwen3-30B-A3B-Base & Base & 30B & Jul 2025\\
      & Qwen3-30B-A3B-Instruct-2507 & Instruct & 30B & Jul 2025 \\
      & Qwen3-30B-A3B-Thinking-2507 & Reasoning & 30B & Jul 2025 \\
    \hline
    \multirow{2}{*}{DeepSeek AI} 
      & DeepSeek-R1-Distill-Llama-70B & Reasoning & 70B & Jan 2025 \\
      \cline{2-5}
      & DeepSeek-R1-Distill-Qwen-32B & Reasoning & 32B & Jun 2024 \\
    \hline
    \multirow{2}{*}{Google} 
      & gemma-3-27b-pt & Base & 27B & Mar 2025\\
      & gemma-3-27b-it & Instruct & 27B & Mar 2025\\
    \hline
  \end{tabular}
  \caption[LLM Selection]{\textbf{Overview of LLMs used in the experiments of this paper by provider, type, size, and release date.} Model families of corresponding base, instruct and reasoning variants have been grouped together.}
  \label{tab:updated-llm-selection}
\end{table*}

\label{app_tab:model-selection-table}
\begin{table*}[t]
\centering
\renewcommand{\arraystretch}{1.15}
\begin{tabular}{l l l}
\hline
\textbf{Provider} & \textbf{Model Family} & \textbf{Reference} \\
\hline

OpenAI & GPT-OSS & \cite{openai_gpt_oss_2025} \\
\hline

Mistral AI & NeMo & \cite{mistral_nemo_2024} \\
& Ministral & \cite{mistral_ministral_2024} \\
& Mistral Small 3 & \cite{mistral_small3_2025} \\
& Magistral & \cite{mistral_magistral_2025} \\
\hline

Meta & LLaMA 3 & \cite{llama3_2024} \\
& LLaMA 4 & \cite{llama4_herd_2025} \\
\hline

Qwen & Qwen3 & \cite{qwen3_2025} \\
\hline

DeepSeek AI & DeepSeek-R1 & \cite{deepseek_r1_2025} \\
\hline

Google DeepMind & Gemma 3 & \cite{gemma3_2025} \\
\hline

\end{tabular}
\caption{\textbf{Model families used in this study with primary technical references.} We group base, instruction-tuned, and reasoning variants under their respective model families.}
\label{tab:model_families}
\end{table*}
\label{app_tab:model-citations}

Please note that the model \model{Magistral-Small-2507} requires using a specific system prompt to enable reasoning\footnote{\url{https://huggingface.co/mistralai/Magistral-Small-2507}}. We have tested the model without a system prompt (acting as an instruction tuned model) and with the system prompt provided, acting as a reasoning model. To highlight whether the reasoning behaviour was enabled through system prompt usage, we added a "Reasoning-Enabled" postfix to the model name in tables and plots.

\subsubsection{Model Configuration and Exceptions}
All models were evaluated in their default configurations. No system prompts were employed, with the sole exception of \model{Magistral-Small-2507}, which requires a system prompt to enable reasoning prior to generating a final answer\footnote{see \href{https://huggingface.co/mistralai/Magistral-Small-2506}{https://huggingface.co/mistralai/Magistral-Small-2506}}. Without the system prompt, the model will behave like an instruction-tuned model. For the label-probability experiments, this model was tested in two variants: with reasoning enabled (\model{Magistral-Small-2507-Reasoning-Enabled}) and without reasoning (\model{Magistral-Small-2507}).

The base (pre-trained) model \model{gemma-3-27b-pt}, corresponding to the instruction-tuned \model{gemma-3-27b-it}, was excluded from the label-probability calibration experiments. In preliminary evaluations, the base model exhibited insufficient task comprehension, resulting in negligible probability mass assigned to label tokens within the top-20 most probable tokens. Consequently, label probabilities could not be retrieved through \texttt{vllm}, preventing the computation of confidence scores.

All models were used in their base configuration. With the exception of \model{Magistral-Small-2507}, no system prompts were used. \model{Magistral-Small-2507} uses the system prompt to elicit reasoning steps before providing a final answer. It was included in the label probability experiment both with (\model{Magistral-Small-2507-Reasoning-Enabled}) and without (\model{Magistral-Small-2507}) the system prompt.
\model{gemma-3-27b-pt}, the pre-trained or base variant of \model{gemma-3-27b-it} was excluded from the experiment researching the label probability calibration due to lack in task comprehension. This resulted in no probability mass being assigned to label tokens within the 20 most probable tokens, rendering probabilities unaccessible via \texttt{vllm}. Therefore no certainties could be retrieved.

For the exact configuration parameters supplied to \texttt{vllm}, please refer to the \lstinline{models.py} files located in the experiment subdirectories of the repository accompanying this paper.

\subsection{Dataset Selection}
\label{app_ssec:dataset_sel}

The datasets used in the experiments are listed in \Cref{tab:scientific-qa-datasets}.
\begin{table*}[t]
\centering
\renewcommand{\arraystretch}{1.3}
\begin{tabularx}{\linewidth}{l c c c X} 
\hline
\textbf{Dataset} & \textbf{Size} & \textbf{Task Format} & \textbf{License} & \textbf{Domain} \\
\hline
\textbf{MMLU}   & 15,908 & Multiple Choice & MIT & General (57 topics, including Physics) \\
\textbf{ARC}    & 7,787  & Multiple Choice & CC-BY-NC 4.0 & Science (Physics, Chemistry, Biology, Earth Science) \\
\textbf{SciQ}   & 13,679 & Multiple Choice & CC-BY-SA 3.0 & Science (Physics, Chemistry, Biology) \\
\textbf{GPQA}   & 448    & Multiple Choice & MIT & Science (Physics, Chemistry, Biology) -- Graduate-level \\
\textbf{GSM8K}  & 8,792  & Arithmetic      & MIT & Mathematics \\
\textbf{GSM-MC} & 8,787  & Multiple Choice & MIT & Mathematics \\
\textbf{SVAMP}  & 1,000  & Arithmetic      & CC-BY 4.0 & Mathematics \\
\textbf{SciBench} & 2,229 & Arithmetic     & CC-BY-SA 4.0 & Science (Physics, Chemistry, Biology, Medicine, Earth Science) \\
\hline
\end{tabularx}
\caption[Survey of Scientific Question Answering Datasets]{%
  \textbf{Scientific QA datasets surveyed.} Selection criteria emphasized natural-science domains (especially physics) and inclusion of different reasoning requirements for answering, while providing a verifiable ground-truth.
}
\label{tab:scientific-qa-datasets}
\end{table*}

\label{datasets-selection-table}
For the exact dataset configuration parameters, please refer to the \path{data_source} subdirectories located in the experiment directories of the repository accompanying this paper.

\subsection{Prompt Composition}
\label{appendix:prompt_composition}
The second experiment evaluates a total of \samplecount{} long-form responses across models and datasets using sequence-level UQ methods. We include five multiple-choice QA datasets (MMLU, ARC-Easy, ARC-Reasoning, SciQ, GPQA) and three arithmetic QA datasets (GSM8K, SVAMP, SciBench).

For the MCQA datasets, we apply counterfactual prompting using the APriCoT strategy, evaluating each of the four answer options independently. This expands each item into four separate prompts. All datasets are subsampled to 250 items, and we sample 10 responses per prompt. Consequently, each arithmetic dataset yields 2,500 responses per model, and each MCQA dataset yields 10,000 responses per model, totaling 57,500 responses per model.

Due to high inference costs, we reduced the number of generated responses specifically for the \model{deepseek-ai/DeepSeek-R1-Distill-Llama-70B} model on the GPQA dataset by half, resulting in 5,000 evaluated responses for this model–dataset pair.

\begin{table*}[t]
\centering
\small
\setlength{\tabcolsep}{2pt}
\renewcommand{\arraystretch}{1.1}
\begin{tabular}{l *{8}{c} r}
\toprule
\textbf{Model} & \rotatebox{90}{\textbf{MMLU}} & \rotatebox{90}{\textbf{ARC-Easy}} & \rotatebox{90}{\textbf{ARC-Reasoning}} & \rotatebox{90}{\textbf{SciQ}} & \rotatebox{90}{\textbf{GPQA}} & \rotatebox{90}{\textbf{GSM8K}} & \rotatebox{90}{\textbf{SVAMP}} & \rotatebox{90}{\textbf{SciBench}} & \textbf{Total} \\
\midrule
gpt-oss-20b & 10k & 10k & 10k & 10k & 10k & 2.5k & 2.5k & 2.5k & 57.5k \\
gpt-oss-120b & 10k & 10k & 10k & 10k & 10k & 2.5k & 2.5k & 2.5k & 57.5k \\
Ministral-8B-Instruct-2410 & 10k & 10k & 10k & 10k & 10k & 2.5k & 2.5k & 2.5k & 57.5k \\
Mistral-Nemo-7B & 10k & 10k & 10k & 10k & 10k & 2.5k & 2.5k & 2.5k & 57.5k \\
Mistral-Small-3.2-24B-Instruct-2506 & 10k & 10k & 10k & 10k & 10k & 2.5k & 2.5k & 2.5k & 57.5k \\
Llama-3.3-70B-Instruct & 10k & 10k & 10k & 10k & 10k & 2.5k & 2.5k & 2.5k & 57.5k \\
Llama-4-Scout-17B-16E-Instruct & 10k & 10k & 10k & 10k & 10k & 2.5k & 2.5k & 2.5k & 57.5k \\
gemma-3-27b-it & 10k & 10k & 10k & 10k & 10k & 2.5k & 2.5k & 2.5k & 57.5k \\
Qwen3-30B-A3B-Instruct-2507 & 10k & 10k & 10k & 10k & 10k & 2.5k & 2.5k & 2.5k & 57.5k \\
Qwen3-30B-A3B-Thinking-2507 & 10k & 10k & 10k & 10k & 10k & 2.5k & 2.5k & 2.5k & 57.5k \\
\midrule
DeepSeek-R1-Distill-Qwen-32B & 10k & 10k & 10k & 10k & 10k & 2.5k & 2.5k & 2.5k & 57.5k \\
DeepSeek-R1-Distill-Llama-70B & 10k & 10k & 10k & 10k & 5k & 2.5k & 2.5k & 2.5k & 52.5k \\
\midrule
\textbf{Total} & 120k & 120k & 120k & 120k & 115k & 30k & 30k & 30k & \textbf{685k} \\
\bottomrule
\end{tabular}
\caption{\textbf{Response counts for Experiment 2 per model and dataset.} Values are in thousands of responses (k). Each MCQA dataset contributes 10{,}000 responses per model (250 items $\times$ 4 APriCoT option prompts $\times$ 10 generations) and each arithmetic dataset 2{,}500 responses per model (250 items $\times$ 10 generations), yielding 57{,}500 responses per model. The sole exception is \model{DeepSeek-R1-Distill-Llama-70B} on GPQA, for which 5{,}000 responses were generated due to high inference costs. This table is the authoritative accounting of the \samplecount{} responses evaluated in Experiment 2.}
\label{tab:response_counts_exp2}
\end{table*}

Across all 12 evaluated models, this results in a total of \samplecount{} responses, each assessed individually using the four selected UQ methods in Experiment 2. The authoritative per-model, per-dataset accounting is provided in \Cref{tab:response_counts_exp2}.

\subsection{Relevance of the Selected Methods to the Field of UQ in LLMs}
\label{app_ssec:method_citation}

\begin{table*}[t]
\centering
\begin{tabular}{lc}
\hline
\textbf{Paper} & \textbf{Citation Count} \\ \hline
Verbalized Uncertainty \citep{2305.14975} & $602$ \\ 
P(True) \citep{2207.05221} & $723$ \\ 
CCP \citep{2403.04696} & $118$ \\ 
Semantic consistency (\citet{2203.11171}, basis of Answer Consistency) & $3831$\\ 
\hline
\end{tabular}
\caption{\textbf{Citation Counts of employed UQ methods (as of writing).}}
\label{tab:uq_citation_counts}
\end{table*}

Obtained from Google Scholar at the time of writing (see \Cref{tab:uq_citation_counts}), Verbalized Uncertainty (as proposed by \citep{2305.14975}) has been cited $602$ times, P(True) (as proposed by \citep{2207.05221} ) has been cited 723 times and CCP (as proposed by \citep{2403.04696}) has been cited 118 times, highlighting their relevance to the field. While Answer Consistency methods have not been directly proposed, it captures semantic consistency and proxies other approaches like self-consistency prompting as detailed in \citep{2203.11171}, which has been cited 3831 times. We found different citation counts on OpenAlex \footnote{\href{https://openalex.org}{openalex.org}} for entries in \Cref{tab:uq_citation_counts} but their relative scale is equivalent than reported here.

To underpin the relevance of the four selected UQ methods, a glimpse at the derivative work is instructive. For Verbalized Uncertainty \citep{2305.14975}, the following papers have built on this idea: \citet{2306.13063, zhao-etal-2024-fact, zhou2025steerconfsteeringllmsconfidence, shrivastava2023llamasknowgptsdont, han2024enhancingconfidenceexpressionlarge}. The idea of P(True) was picked up by \citet{Mahaut_2024,
chen-mueller-2024-quantifying,
ren2023selfevaluationimprovesselectivegeneration,
zhang2025cotuqimprovingresponsewiseuncertainty,
xiao2025enhancinguncertaintyestimationllms,
heo2025llmsestimateuncertaintyinstructionfollowing,
kapoor2025largelanguagemodelstaught}. This demonstrates how essential a rigorous study of the effectiveness and performance of these methods is.

Beyond the concrete P(True) method itself, the broader problem of whether token probabilities constitute reliable confidence estimates is highly relevant to LLM-as-a-Judge systems and classifier-style LLM applications in general. Many such systems implicitly rely on token probabilities as confidence signals for labels or decisions, making their calibration directly relevant for the reliability of automated evaluation and downstream decision-making pipelines.


\subsection{Uncertainty Quantification Method Exclusion}
\label{app_ssec:unc_method_exclusion}

Our selection of UQ methods reflects principled constraints: feasibility, interpretability, and applicability to realistic scientific reasoning tasks. An overview over exclusion of methods and short form explanation is provided in \Cref{tab:exclusion-of-metrics}. Only methods producing normalized sequence-level uncertainty scores are included in our analysis to enable reliability UQ validation by calibration. Subsequently, we focus our work on the following UQ methods: \textit{Verbalized Uncertainty} \citep{2305.14975}, \textit{P(True)} \citep{2207.05221}, \textit{Answer Consistency} \citep{2203.11171}, \textit{Claim-Conditioned Probability (CCP)} \citep{2403.04696}. More details of these methods are discussed in \Cref{sec:metric_implementation}.

\subsubsection{Survey of UQ methods}
We used the taxonomy from \texttt{LM-Polygraph} \citet{fadeeva-etal-2023-lm} as a starting point and adapted the categorization of UQ methods into information-based, ensemble-based, density-based, reflexive or verbalized and meaning-diversity approaches. However, it is important to emphasize that \texttt{LM-Polygraph} targets selective generation, whereas our work focuses on calibration, the second core UQ task. Calibration requires normalized confidence scores in [0,1], which immediately excludes many of the 28 methods surveyed in \texttt{LM-Polygraph}. Thus, while we relied on \citet{fadeeva-etal-2023-lm} for a broad overview of UQ approaches, the spectrum of methods applicable to calibration is necessarily narrower.

\subsubsection{Density-Based Methods}
Density-based methods require access to the model’s training data to estimate likelihoods or information-theoretic quantities. In our benchmark, the evaluated models are open-weight but not open-source, and their training corpora are unavailable. Consequently, density-based approaches cannot be applied, making them unsuitable for our calibration-focused evaluation.

\subsubsection{Ensemble-Based Methods}
Ensemble methods combine multiple distinct models to estimate uncertainty. While potentially effective, they prevent the assessment of per-model calibration, which is central to our study. Additionally, ensembles introduce substantial design ambiguity regarding which combinations of models to use and incur significant computational costs. In contrast, the Answer Consistency method we benchmark samples a single model multiple times, preserving per-model interpretability while allowing assessment of semantic consensus as an UQ signal.

\subsubsection{Claim-Level Methods}
Semantic Entropy as a claim-level method was excluded due to methodological and computational limitations. Existing evaluations of this method are restricted to short-form factual QA datasets and do not assess calibration, offering limited evidence for their suitability in reasoning-intensive scientific QA. Furthermore, scientific answers frequently span hundreds or thousands of tokens. In our internal tests, claim-extraction prompts often produced tens to hundreds of trivial or irrelevant claims, each requiring separate LLM evaluation. This leads to computational costs far exceeding those of the  Answer Consistency method, our computationally most expensive method employed, making claim-level approaches impractical for large-scale scientific QA.

\subsubsection{Overview Over Excluded Methods}
The full list of excluded methods with their exclusion reasons is provided in \Cref{tab:exclusion-of-metrics}.
\clearpage
\onecolumn
\footnotesize
\begin{longtable}[c]{@{}p{6cm}p{3cm}p{6.2cm}@{}}
\toprule
\textbf{Method} & \textbf{Category} & \textbf{Exclusion Reasons} \\
\midrule
\endfirsthead

\toprule
\textbf{Method} & \textbf{Category} & \textbf{Exclusion Reasons} \\
\midrule
\endhead

Perplexity \citep{fomicheva-etal-2020-unsupervised}
& Information-based
& Produces unnormalized scores\\
\hline

Mean/max token entropy \citep{fomicheva-etal-2020-unsupervised}
& Information-based
& Produces unnormalized scores\\
\hline

Monte Carlo sequence entropy \citep{2302.09664}
& Information-based
& Produces unnormalized scores\\
\hline

Pointwise mutual information (PMI) \citep{takayama-arase-2019-relevant}
& Information-based
& Produces unnormalized scores\\
\hline

Conditional PMI \citep{van-der-poel-etal-2022-mutual}
& Information-based
& Produces unnormalized scores\\
\hline

Rényi divergence \citep{darrin-etal-2023-rainproof}
& Information-based
& Produces unnormalized scores\\
\hline

Fisher-Rao distance \citep{darrin-etal-2023-rainproof}
& Information-based
& Produces unnormalized scores\\
\hline

Focus \citep{zhang-etal-2023-enhancing-uncertainty}
& Information-based
& Produces unnormalized scores\\
\hline

Semantic entropy \citep{2302.09664}
& Meaning diversity
& Designed to work at the individual claim level rather than on entire sequences\newline Very high computational cost\\
\hline


TokenSAR \citep{2307.01379}
& Meaning diversity
& Alters sentences in a way that violates autoregressive assumptions\newline Relies on NLI models whose performance in scientific contexts is inadequate\\
\hline

SentenceSAR \citep{2307.01379}
& Meaning diversity
& Alters sentences in a way that violates autoregressive assumptions\newline Relies on NLI models whose performance in scientific contexts is inadequate\\
\hline

SAR \citep{2307.01379}
& Meaning diversity
& Alters sentences in a way that violates autoregressive assumptions\newline Relies on NLI models whose performance in scientific contexts is inadequate\\
\hline

EigenScore \citep{chen2024inside}
& Meaning diversity
& Produces unnormalized scores\\
\hline

Sentence-level ensemble-based measures \citep{malinin2020uncertainty}
& Ensembling
& Requires running multiple independent models, leading to high computational cost\newline Introduces extra variability that complicates comparison to single-model methods\\
\hline

Token-level ensemble-based measures \citep{malinin2020uncertainty}
& Ensembling
& Requires running multiple independent models, leading to high computational cost\newline Introduces extra variability that complicates comparison to single-model methods\\
\hline

Mahalanobis distance (MD) \citep{lee2018simple}
& Density-based
& Density-based approach that requires proprietary training data\\
\hline

Robust density estimation (RDE) \citep{yoo-etal-2022-detection}
& Density-based
& Density-based approach that requires proprietary training data\\
\hline

Relative Mahalanobis distance (RMD) \citep{ren2023out}
& Density-based
& Density-based approach that requires proprietary training data\\
\hline

Hybrid Uncertainty Quantification (HUQ) \citep{vazhentsev-etal-2023-hybrid}
& Density-based
& Density-based approach that requires proprietary training data\\
\hline

Number of semantic sets (NumSets) \citep{lin2023generating}
& Meaning Diversity
& Produces count-based outputs that are not normalized to [0,1]\\
\hline

Sum of eigenvalues of the graph Laplacian (EigV) \citep{lin2023generating}
& Meaning Diversity
& Produces unnormalized scores\\
\hline

Degree matrix (Deg) \citep{lin2023generating}
& Meaning Diversity
& Produces unnormalized scores\\
\hline

Eccentricity (Ecc) \citep{lin2023generating}
& Meaning Diversity
& Produces unnormalized scores\\
\hline

Lexical similarity (LexSim) \citep{fomicheva-etal-2020-unsupervised}
& Meaning Diversity
& Relies on surface-level token overlap instead of semantic meaning\\
\hline

Kernel Language Entropy \citep{nikitin2024kernel}
& Meaning Diversity
& Produces unnormalized scores\\
\hline

LUQ (\citep{zhang-etal-2024-luq})
& Meaning diversity
& Produces unnormalized scores\\
\hline

\bottomrule

\caption{\textbf{Reasons for Exclusion of Uncertainty Metrics in Benchmark.} List of UQ Methods was adapted from the comprehensive evaluation and classification of uncertainty methods by \citet{fadeeva-etal-2023-lm}.}
\label{tab:exclusion-of-metrics}\\
\end{longtable}
\normalsize
\clearpage
\twocolumn

\subsection{Label Probability Calibration Experiment Design}
\label{sec:label_prob_exp_design_appendix}
For the Label Probability Calibration Experiment, preliminary tests revealed that base models struggled with task comprehension: in many cases, none of the valid label tokens (A/B/C/D) appeared among the top-20 first-token predictions. To address this, we designed four prompt variants with different structural formats to identify the formulation that maximizes task comprehension. We define task comprehension as the average total probability mass assigned to the label tokens (A/B/C/D) in the model’s first generated token.

The four prompt designs are presented in \Cref{sec:exp1_prompt_designs}. An overview of task comprehension scores for each prompt is provided in \Cref{sec:task_comprehension_prompts}, where Prompt 1 emerges as the most effective. Nevertheless, even with improved task comprehension, a substantial fraction of responses remained invalid (i.e. no extractable probability assigned to any label), as detailed in \Cref{sec:high_invalid_answer_count}.

We analyze the causes of these invalid responses in \Cref{sec:invalid_answer_analysis} and outline the mitigation strategies applied to reduce them in \Cref{sec:invalid_answer_mitigation}. Finally, we report the updated invalid-answer statistics for Prompt 1 after applying these adjustments in \Cref{sec:high_invalid_answer_count}. We report the distribution of token counts for the reasoning chains for reasoning models in \Cref{sec:exp1_reasoning_lengths}.
\subsubsection{Prompt Designs}
\label{sec:exp1_prompt_designs}
\begin{lstlisting}[
  style=plaintextfile,
  captionpos=t,
  caption = {[Prompt Design 1 for Experiment 1]\textbf{Prompt Design 1 for Experiment 1.} The prompt design uses 3-shot prompting. \texttt{<QUESTION>} and \texttt{<ANSWER CHOICE X>} are replaced with the individual questions and answer choices from the benchmarked dataset. Prompt Design 1 features markdown construct.},
]
You are a highly capable language model trained for multiple-choice question answering.
Below are three examples of multiple-choice questions with labeled answer choices. Each example includes the correct answer.
After the examples, you will be given a new question with four labeled answer choices (A, B, C, D).

Your task is to select the answer choice you believe is correct by responding with only the corresponding label: A, B, C, or D.
Do not include any explanation or additional text.

### Example 1:
**Question:** What is the capital of France?
A) Berlin
B) Madrid
C) Paris
D) Rome

**Correct Answer:** C

<Two More Examples Omitted for Readability>

---

**Question:** <QUESTION>
A) <ANSWER CHOICE A>
B) <ANSWER CHOICE B>
C) <ANSWER CHOICE C>
D) <ANSWER CHOICE D>

**Correct Answer:** 
\end{lstlisting}

\begin{lstlisting}[
  style=plaintextfile,
  captionpos=t,
  caption = {[Prompt Design 2 for Experiment 1]\textbf{Prompt Design 2 for Experiment 1.} The prompt design uses 3-shot prompting. \texttt{<QUESTION>} and \texttt{<ANSWER CHOICE X>} are replaced with the individual questions and answer choices from the benchmarked dataset. Prompt Design 2 features no introductary text, role or task description. The format is designed to represent natural language without special formatting.},
]
Question: What is the capital of France?
A) Berlin
B) Madrid
C) Paris
D) Rome

The correct answer is C

<Two More Examples Omitted for Readability>

Question: <QUESTION>
A) <ANSWER CHOICE A>
B) <ANSWER CHOICE B>
C) <ANSWER CHOICE C>
D) <ANSWER CHOICE D>

The correct answer is 
\end{lstlisting}

\begin{lstlisting}[
  style=plaintextfile,
  captionpos=t,
  caption = {[Prompt Design 3 for Experiment 1]\textbf{Prompt Design 3 for Experiment 1.} The prompt design uses 3-shot prompting. \texttt{<QUESTION>} and \texttt{<ANSWER CHOICE X>} are replaced with the individual questions and answer choices from the benchmarked dataset. The format of Prompt 3 is designed to represent natural language without special formatting. The format of the answer specifically requests the label, not the answer in general.},
]
You are a highly capable language model trained for multiple-choice question answering. In the following examples, you will see questions with answer choices. The answer choices are preceded by the phrase "Answer Choices:". Each answer choice is annotated with one of the labels A, B, C or D. The correct answer to the question is given by the sentence "The label of the correct answer choice is" followed by the corresponding label. Your task is to answer the new question in the same format, outputting only the label of the correct answer to the question you are provided. Do not output anything other than one of the labels A, B, C or D.

Question: What is the capital of France?
Answer Choices:
A) Berlin
B) Madrid
C) Paris
D) Rome

The label of the correct answer choice is C

<Two More Examples Omitted for Readability>

Question: <QUESTION>
Answer Choices:
A) <ANSWER CHOICE A>
B) <ANSWER CHOICE B>
C) <ANSWER CHOICE C>
D) <ANSWER CHOICE D>

The label of the correct answer choice is 
\end{lstlisting}

\begin{lstlisting}[
  style=plaintextfile,
  captionpos=t,
  caption = {[Prompt Design 4 for Experiment 1]\textbf{Prompt Design 4 for Experiment 1.} The prompt design uses 3-shot prompting. \texttt{<QUESTION>} and \texttt{<ANSWER CHOICE X>} are replaced with the individual questions and answer choices from the benchmarked dataset. Prompt Design 2 features special tags to mark the answer given as a label.},
]
You are a highly capable multiple-choice question answering model. Below are three examples that show the format you must follow. Each question has four answer choices labeled A, B, C, and D. Your task is to answer a new question by outputting the correct answer in the following format: <ANSWER>X<ANSWER>, where X is the label corresponding to the correct answer, A, B, C or D. Do not add any extra text or explanation.

Example 1:
Question: What is the capital of France?
Answer Choices:
A) Berlin
B) Madrid
C) Paris
D) Rome

<ANSWER>C<ANSWER>

<Two More Examples Omitted for Readability>

Now, please answer the following question in the same format.

Question: <QUESTION>
Answer Choices:
A) <ANSWER CHOICE A>
B) <ANSWER CHOICE B>
C) <ANSWER CHOICE C>
D) <ANSWER CHOICE D>

<ANSWER>
\end{lstlisting}

\subsubsection{Task Comprehension per Prompt Design}
\label{sec:task_comprehension_prompts}
\begin{table*}[t]
\centering
\begin{tabularx}{\textwidth}{lrrrr}
\toprule
\textbf{Model Category} & \textbf{Prompt 1} & \textbf{Prompt 2} & \textbf{Prompt 3} & \textbf{Prompt 4} \\
\midrule
Base Models              & 0.2368 & 0.5928 & 0.4135 & 0.1632 \\
Instruction-Tuned Models & 0.9912 & 0.3597 & 0.9561 & 0.2646 \\
Reasoning Models         & 0.7002 & 0.0895 & 0.4475 & 0.0022 \\
\midrule
\textbf{Average}         & \textbf{0.6427} & \textbf{0.3474} & \textbf{0.6057} & \textbf{0.1434} \\
\bottomrule
\end{tabularx}
\caption{\textbf{Task Comprehension Measured by Probability Mass Assigned to Answer Labels.} 
Mean over the sum of label probabilities per question for base, instruction-tuned, and reasoning models. 
On average, task comprehension is highest under Prompt~1.}
\label{tab:labelprobsum}
\end{table*}

\subsubsection{Invalid Answer Count}
\label{sec:high_invalid_answer_count}
Although we selected the prompt that yielded the highest task comprehension for our benchmark, the original experiment exhibited a substantial number of invalid answers, varying with prompt design. These invalid outputs contaminated the results. After performing analysis on the cause of this behaviour and employing mitigation strategies, we were able to significantly decrease the amount of invalid answers, as detailed in \Cref{tab:invalid_answers}.
\begin{table*}[t]
\centering
\small
\begin{tabularx}{\textwidth}{lrrrr|r}
\toprule
\textbf{Model} & \textbf{Prompt 1} & \textbf{Prompt 2} & \textbf{Prompt 3} & \textbf{Prompt 4} & \textbf{Prompt 1 (SD)}\\
\midrule
gpt-oss-20b & 307  & 3356  & 498   & 25304 & 395 \\
gpt-oss-120b & 38  & 3323  & 27    & 24873 & 0 \\
Ministral-8B-Instruct-2410 & 0    & 0     & 0     & 0 & 0 \\
Mistral-Nemo-Base-2407 & 0        & 0     & 12574 & 0 & 0 \\
Mistral-Nemo-Instruct-2407 & 0     & 0     & 0     & 0 & 0 \\
Mistral-Small-3.1-24B-Base-2503 & 0 & 0     & 0     & 0 & 0 \\
\makecell[tl]{Mistral-Small-3.2-24B-\\\hspace*{1cm}Instruct-2506} & 0 & 0 & 0     & 0 & 0 \\
Magistral-Small-2507 & 0          & 0     & 0     & 0 & 0 \\
\makecell[tl]{Magistral-Small-2507-\\\hspace*{1cm}Reasoning-Enabled} & 8604 & 18716 & 4749 & 21232 & 123 \\
Llama-3.1-70B & 0                 & 13    & 0     & 6636 & 0 \\
Llama-3.3-70B-Instruct & 0        & 2527  & 19    & 125 & 0 \\
Llama-4-Scout-17B-16E & 2         & 3     & 0     & 5048 & 0 \\
Llama-4-Scout-17B-16E-Instruct & 0 & 15   & 0     & 0 & 0 \\
Qwen3-30B-A3B-Base & 2           & 1     & 0     & 55 & 0 \\
Qwen3-30B-A3B-Instruct-2507 & 0   & 6270  & 0     & 722 & 0 \\
Qwen3-30B-A3B-Thinking-2507 & 6242 & 11746 & 8656 & 21840 & 58 \\
DeepSeek-R1-Distill-Llama-70B & 425 & 7538 & 725  & 7477 & 148 \\
DeepSeek-R1-Distill-Qwen-32B & 914 & 7465 & 1170 & 3337 & 243 \\
gemma-3-27b-pt & 25316 & 25316 & 25316 & 25316 & 0 \\
gemma-3-27b-it & 0               & 431   & 0     & 2 & 0 \\
\bottomrule
\end{tabularx}
\caption{\textbf{Number of Invalid Answers Given by Models for Different Prompt Designs Across All Datasets (\(n = 25{,}316\)).} Invalid answers assign no probability mass to any answer-choice labels. This phenomenon is discussed above. Using the prompt design yielding the best task comprehension (as discussed in \Cref{sec:task_comprehension_prompts}), we adjusted the experiment to use structured decoding and increased the maximum token generation limit. After this adjustment, invalid answers may only occur when reasoning chains exceed 10240 tokens. The result is shown in column "Prompt 1 (SD)", drastically decreasing the amount of invalid answers.}
\label{tab:invalid_answers}
\end{table*}

\subsubsection{Analysis of Invalid Answers}
\label{sec:invalid_answer_analysis}
Our investigation identified two primary causes of invalid responses:

\begin{enumerate}
    \item \textbf{Missing answer labels within top token alternatives:} Models sometimes did not generate any of the expected labels (A/B/C/D) within the top 20 token alternatives that \texttt{vLLM} provides per token. This occurred in:
    \begin{enumerate}
        \item \emph{Base models}, which lacked sufficient task comprehension and failed to include any of the answer labels in the top 20 token alternatives.
        \item \emph{Reasoning models}, which faced the additional challenge of adhering to the expected output format after generating long-form reasoning, as the attention mechanism favours more recent tokens. Reasoning of several thousand tokens appeared to negatively effect task comprehension, leading to invalid answers in extreme cases.
    \end{enumerate}

    \item \textbf{Overflowing generations:} Reasoning models produced outputs exceeding the originally set maximum token limit (4096 tokens), appearing to be stuck in prolonged reasoning cycles without converging to a final answer.
\end{enumerate}

A special case is \model{Magistral-Small-2507}, which has reasoning enabled via the system prompt provided by MistralAI\footnote{\url{https://huggingface.co/mistralai/Magistral-Small-2506}}
. For this model, we consistently observed non-converging generations that produced a very high number of invalid answers, even on the relatively simple MMLU dataset compared to GPQA. Increasing the maximum token generation did not resolve the issue -- the model still consistently hit the upper generation limit. Our solution for handling these invalid outputs is detailed in \Cref{sec:magistral-invalid-answers}.

\subsubsection{Mitigation Strategies}
\label{sec:invalid_answer_mitigation}

We addressed these issues as follows:

\begin{itemize}
    \item Leveraged \texttt{vLLM}'s structured decoding feature to enforce compliance with the expected output format after reasoning using the regex \texttt{"[ABCD]"}.
    \item Increased the maximum token limit for reasoning from 4096 to 10240 tokens, a 2.5$\times$ increase.
    \item Adjusted the generation of \model{Magistral} to handle infinite generations, as detailed in \Cref{sec:magistral-invalid-answers}.
\end{itemize}

While structured decoding enforces compliance with the expected output format, the prior evaluation of task comprehension per prompt remains essential. Forcing task compliance without adequate comprehension would result in outputs that, although syntactically valid, are semantically incorrect or inconsistent with the question, thereby undermining the reliability of uncertainty quantification and downstream analyses.

Based on the prior evaluation, we chose Prompt 1 for the updated experiment and discarded the other Prompt Designs.

\subsubsection{Magistral-Small-2507 Invalid Answer Solution}
\label{sec:magistral-invalid-answers}
To evaluate \model{Magistral-Small-2507} on label probability calibration and score distribution despite persistent infinite-generation issues, we first set the maximum token limit for reasoning chain generation to 4096 tokens. Generated outputs almost always contained unfinished reasoning chains, indicated by a missing end-of-reasoning token \texttt{"[/THINK]"}. We removed \texttt{"[/THINK]"} if present, trimmed any trailing incomplete sentence, and appended \texttt{". I should now respond with the single label A, B, C, or D associated with the answer I consider most correct.[/THINK]"} to artificially terminate the chain and prompt a single-label response. The model was then queried again with this reasoning chain as a prefix, and label probabilities were extracted from the next generated token. Although structured decoding cannot be applied in this prefix-prompting setup, this procedure reduced invalid answers from 8604 to 123, as shown in \Cref{tab:invalid_answers}.

\subsubsection{Reasoning Token Count Distribution for Reasoning Models}
\label{sec:exp1_reasoning_lengths}
\begin{figure*}[t]
    \centering
     \includegraphics[width=\linewidth]{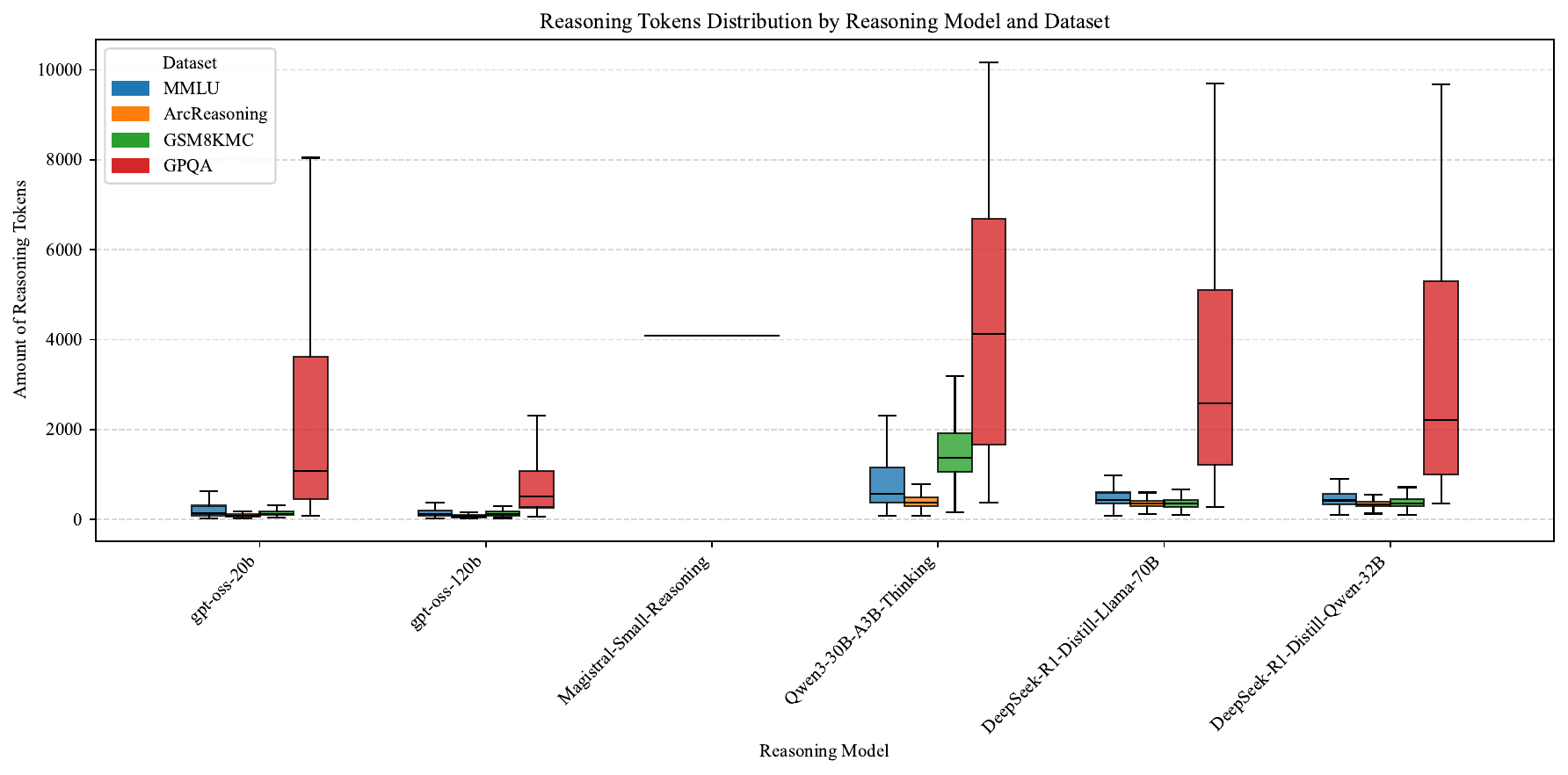}
    \caption{\textbf{Distribution of Reasoning Tokens Produced by Models.} The figure displays a grouped boxplot illustrating the distribution of the Amount of Reasoning Tokens generated by different Reasoning Models across the individual Datasets. Models are grouped along the X-axis, with each boxplot representing the token count distribution for a specific dataset within that model. The central line in each box represents the median, and the box edges represent the first and third quartiles.}
    \label{fig:exp1_reasoning_lengths}
\end{figure*}

The distribution of reasoning token counts per model and dataset is shown in \Cref{fig:exp1_reasoning_lengths}.



\subsection{Label Probability Experiment Results}
\subsubsection{Comprehensive Plots per Prompt with Tables}
\label{sec:exp1_full_plots}
Detailed calibration plots of label probabilities for all prompt designs and configurations are available in the project repository at
\path{label_prob_calibration/resources/figures/full_plots} for the initial prompt designs without structured decoding and at \path{label_prob_calibration/resources_struct_decoding/figures/full_plots} for the chosen Prompt 1 and structured decoding enabled.
Each file follows the naming convention
\path{cal_plot_prompt<id>_table<t>_chosenonly<c>_norm<n>.svg},
where the placeholders encode the following settings:
\begin{itemize}
\item \texttt{prompt}: Identifier of the prompt design used to generate the plot (see \Cref{sec:exp1_prompt_designs}).
\item \texttt{table}: Indicator of whether a table summarizing key calibration statistics is included.
\item \texttt{chosenonly}: Flag specifying whether confidence scores are computed for all candidate labels (\texttt{0}) or restricted to the chosen (most probable) label (\texttt{1}).
\item \texttt{norm}: Flag denoting whether label probabilities are normalized such that their sum equals one across all candidate labels.
\end{itemize}

In the following, the plots showing the calibration plots for Prompt 1 (best task comprehension) can be seen (\Cref{fig:exp1_full_struct_dec,fig:exp1_full_normalized,fig:exp1_full_unnormalized}). The plots are included to visualize systematic patterns in the calibration plots induced by fine-tuning. Full size plots including summary statistics per plot for individual plot assessment can be found in the repository.

\clearpage
\onecolumn
\begin{landscape}
\begin{figure}[H]
    \centering
     \includegraphics[width=\linewidth]{figures/cal_plot_prompt1_struct_dec_table0_chosenonly1_norm1.pdf}
    \caption{\textbf{Calibration Plots for Prompt 1 with structured decoding enabled, using  only the most probable label.}  The columns represent different models, while the rows represent datasets. Tables below the plots list summary metrics such as ECE, normalized entropy of bucket counts and AUROC. Base models are shown in orange, instruction-tuned models in blue, and reasoning models in green. Darker colors indicate a higher number of items in the bin.}
    \label{fig:exp1_full_struct_dec}
\end{figure}

\begin{figure}[H]
    \centering
     \includegraphics[width=\linewidth]{figures/cal_plot_prompt1_table0_chosenonly1_norm1.pdf}
    \caption{\textbf{Calibration Plots for Prompt 1, with structured decoding disabled, using normalized label probabilities and only the most probable label.}  The columns represent different models, while the rows represent datasets. Tables below the plots list summary metrics such as ECE, normalized entropy of bucket counts and AUROC. Base models are shown in orange, instruction-tuned models in blue, and reasoning models in green. Darker colors indicate a higher number of items in the bin.}
    \label{fig:exp1_full_normalized}
\end{figure}

\begin{figure}[H]
    \centering
     \includegraphics[width=\linewidth]{figures/cal_plot_prompt1_table0_chosenonly1_norm0.pdf}
    \caption{\textbf{Calibration Plots for Prompt 1, with structured decoding disabled, using unnormalized label probabilities and only the most probable label.}  The columns represent different models, while the rows represent datasets. Tables below the plots list summary metrics such as ECE, normalized entropy of bucket counts and AUROC. Base models are shown in orange, instruction-tuned models in blue, and reasoning models in green. Darker colors indicate a higher number of items in the bin.}
    \label{fig:exp1_full_unnormalized}
\end{figure}
\end{landscape}
\clearpage
\twocolumn

\subsubsection{Effect of Normalization}
\label{app_sssec:normeff}

\begin{figure*}[t]
    \centering
     \includegraphics[width=\linewidth]{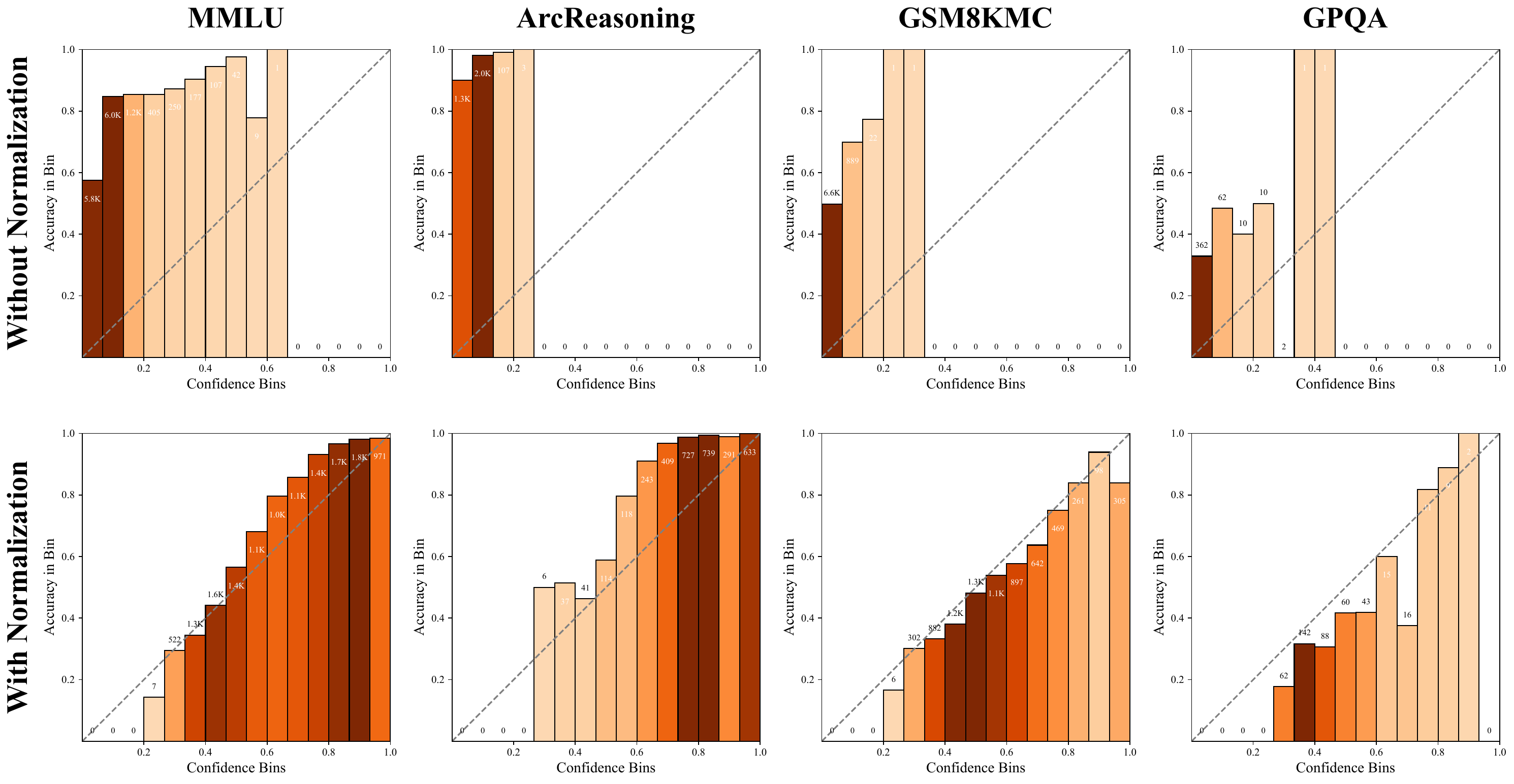}
    \caption[Calibration Plots for Unnormalized and Normalized Label Probabilities]{\textbf{Representative Comparison of Calibration Plots for Unnormalized and Normalized Label Probabilities for the Model \model{Mistral-Small-3.1-24B-Base-2503} and Prompt 1 across all datasets used.}  Calibration improves significantly after normalizing label probabilities.}
    \label{fig:effect_of_normalization}
\end{figure*}

While we had previously hypothesised that using raw label probabilities may enable the model to express general uncertainty, our results show that the task comprehension poses a bigger impact on the total probability mass assigned to label tokens. As a result, we find that raw probabilities are confounded by overall task comprehension and yield misleading calibration, while normalization (excluding non-label tokens) enables meaningful confidence estimates (see \Cref{fig:effect_of_normalization}).

\subsubsection{Selection Bias}
\begin{figure*}[t]
    \centering
     \includegraphics[width=\linewidth]{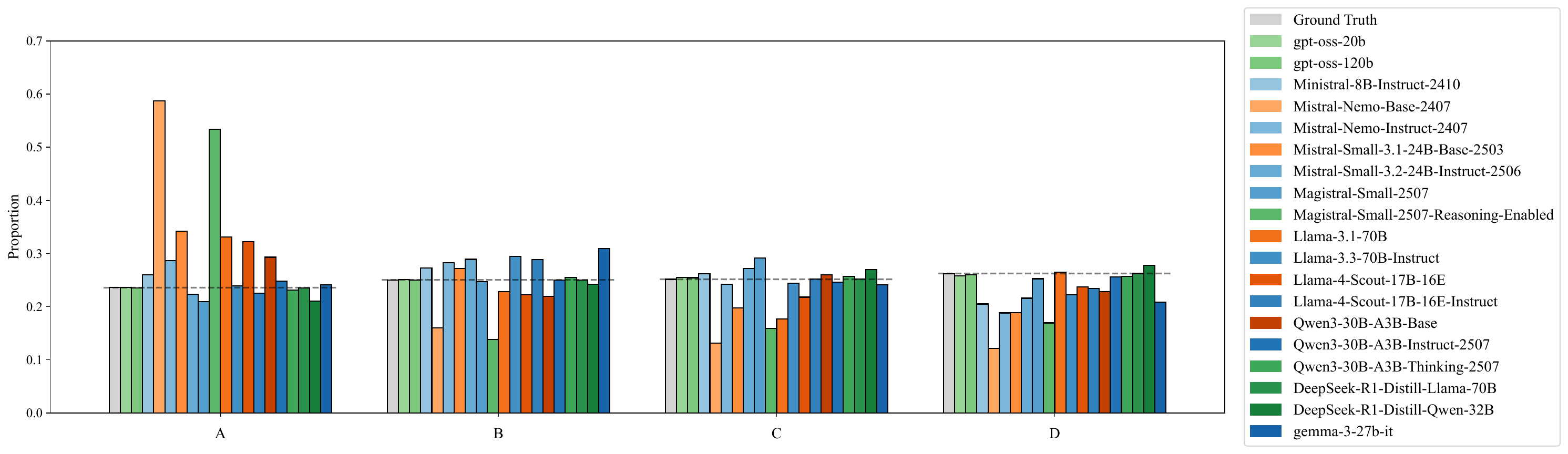}
  \caption[Probabilities for the Labels Summed Across All Datasets for Each Individual Model using Prompt 1]{\textbf{Probabilities for the Labels summed across all Datasets for each individual Model using Prompt 1.}  The ground truth, represented by the distribution of the labels of the correct answers across all items in the datasets, is visualized by the grey bars and the dashed baselines.}
  \label{fig:label_bias_estimations}
\end{figure*}

\Cref{fig:label_bias_estimations} shows the label probabilities aggregated across all datasets for each model under Prompt~1.

\subsubsection{Effect of Finetuning on Uncertainty Score Distributions}
\begin{landscape}
\begin{figure}[H]
    \centering
    \includegraphics[width=\linewidth]{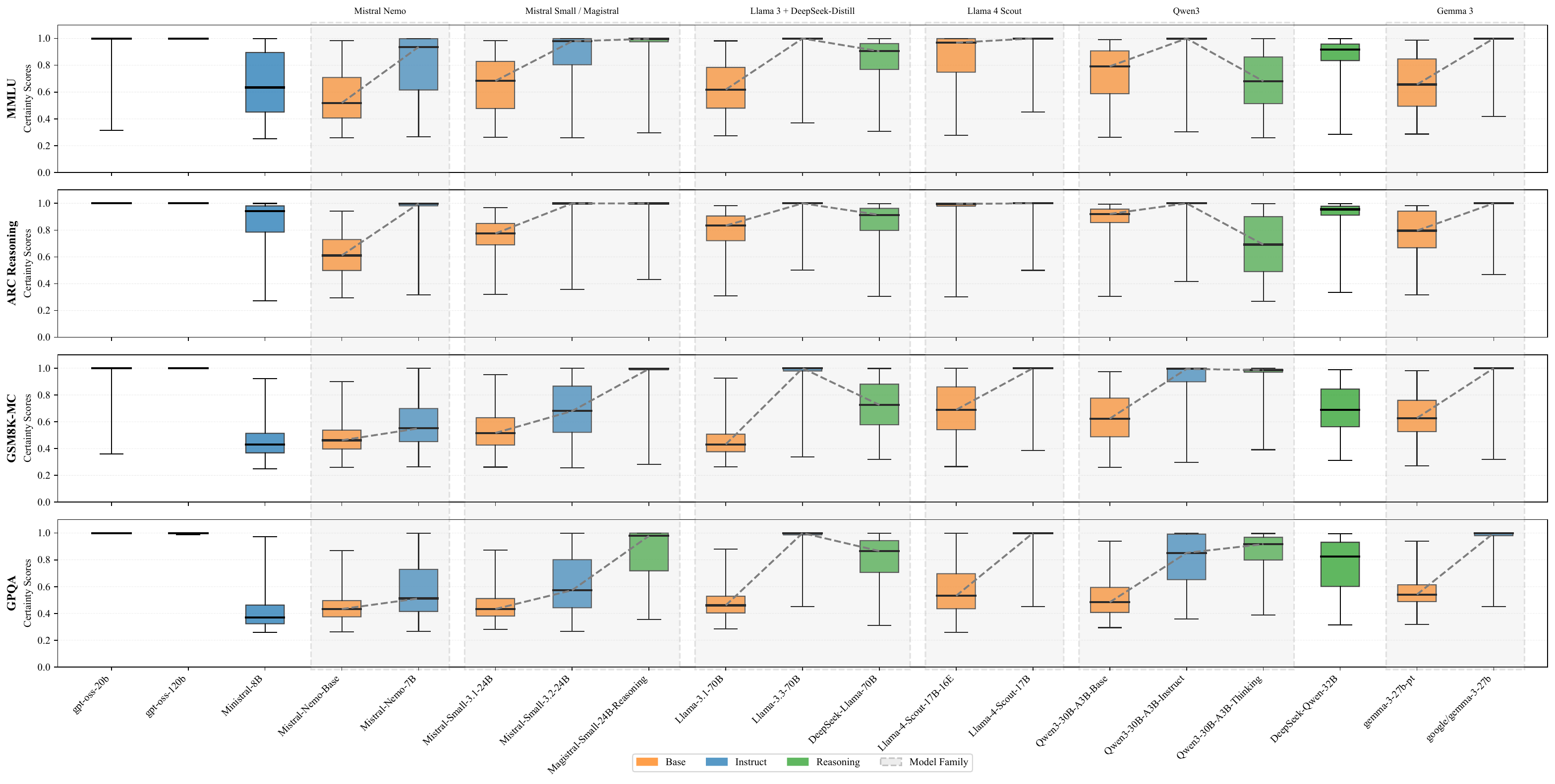}
    \caption[Distribution of normalized certainty scores (chosen option only) across model variants and evaluation datasets]{\textbf{Distribution of normalized certainty scores (chosen option only) across model variants and evaluation datasets.} Rows correspond to datasets and columns to models. Box colors encode model variant types: orange = base models, blue = instruction-tuned models, and green = reasoning models. Light gray dashed rectangles group related models belonging to the same model family across all dataset rows. Within each family, dashed gray lines connect the medians of successive variants (base $\rightarrow$ instruct $\rightarrow$ reasoning), highlighting shifts in certainty distributions induced by fine-tuning. Family names are shown above the grouped columns.}
    \label{fig:polarization_of_score_distributions}
\end{figure}
\end{landscape}

The distribution of normalized certainty scores across model variants and datasets is shown in \Cref{fig:polarization_of_score_distributions}.
\clearpage
\onecolumn
\clearpage
\twocolumn

\subsubsection{Task Comprehension per Prompt Design}
\begin{figure*}[t]
    \centering
     \includegraphics[width=\linewidth]{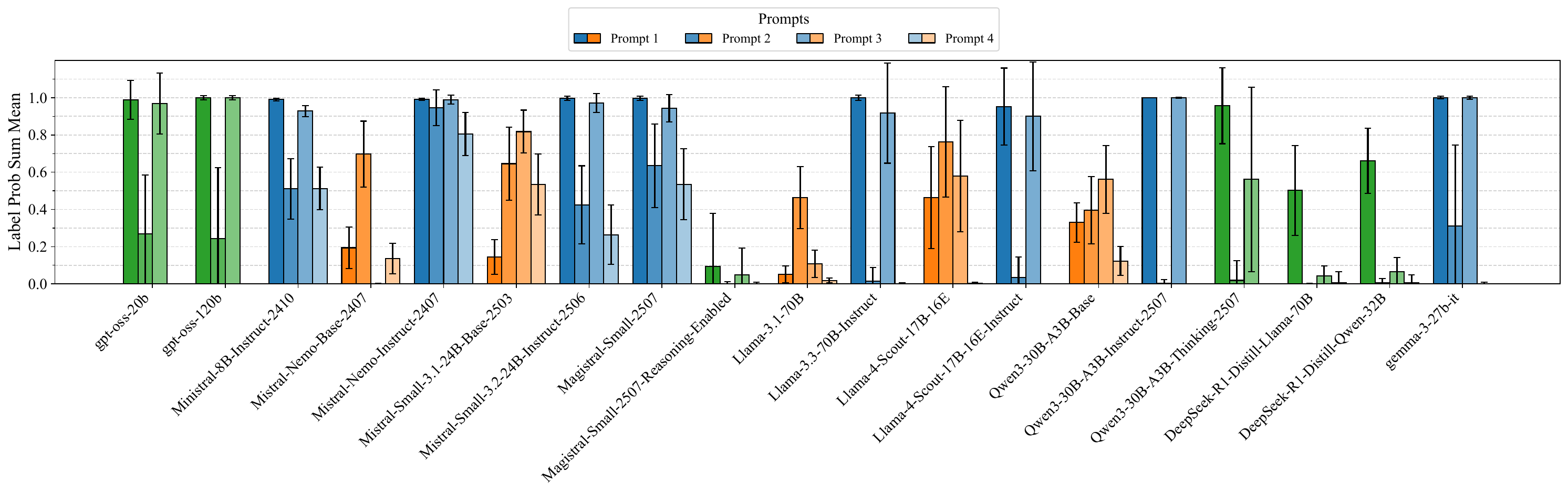}
  \caption[Mean over Sum of Label Probabilities for the Different Prompt Designs and Models.]{\textbf{Task Comprehension per Prompt Design.} Mean over Sum of Label Probabilities for the Different Prompt Designs and Models. The data has been aggregated across all datasets, spanning 25,316 items per model and prompt design. Base models are shown in orange, instruction-tuned models in blue, and reasoning models in green.}
  \label{fig:exp1_task_comprehension_per_prompt}
\end{figure*}

Task comprehension scores per prompt design and model are shown in \Cref{fig:exp1_task_comprehension_per_prompt}.

\subsection{Sequence-Level Calibration}
\subsubsection{Prompts Used For Question Answering}
The sequence-level experiments employed the APriCoT prompting strategy for MCQA and a standard Chain-of-Thought (CoT) approach for arithmetic question answering (Arithmetic QA), followed by final answer extraction.  
The exact prompts for both answer generation and final answer extraction are available in the repository accompanying this paper, specifically in  
\path{/llm-uncertainty-bench/seq_ue_calibration/nodes/apricot_mc_calc.py} for MCQA  
and  
\path{/llm-uncertainty-bench/seq_ue_calibration/nodes/arithmetic_calc.py} for Arithmetic QA.

\subsubsection{Formal Definitions of Answer Consistency}
\label{app_ssec:answer_consistency_definitions}

We provide the formal definitions of the Answer Consistency score for both task types. In both cases, the confidence score for a given sampled generation is computed exclusively from the agreement among the sampled generations; ground-truth (gold) labels are not consulted at any point of this computation. Gold labels are introduced only afterwards, solely to determine whether each model judgment was correct and to compute calibration metrics (\Cref{ssec:eval_metrics}).

\paragraph{Multiple-Choice QA (MCQA).}
Under APriCoT prompting, each MCQA item is evaluated once per answer option. For a given item and answer option $o$, we draw $n=10$ independent generations and obtain a binary judgment $y_i \in \{0,1\}$ for each sampled generation $i \in \{1,\dots,n\}$, where $y_i = 1$ iff generation $i$ concludes that option $o$ is the correct answer. The judgment is extracted from the generation by a follow-up prompt and contains no reference to the gold label. The Answer Consistency score of generation $i$ is the empirical frequency with which the remaining generations agree with its judgment:
\begin{equation}
    \mathrm{AC}_i \;=\; \frac{1}{n-1}\sum_{j \neq i} \mathbb{1}\!\left[\, y_j = y_i \,\right].
\end{equation}
Thus the score is determined entirely by the agreement structure of the sampled judgments $\{y_1,\dots,y_n\}$. The gold label is used only post hoc, to establish the correctness of each judgment $y_i$ for the computation of calibration metrics.

\paragraph{Arithmetic QA.}
For each arithmetic item, we draw $n=10$ independent generations and extract the final numeric answer $\hat{a}_i$ from generation $i$ via a dedicated extraction prompt (for SciBench, semantic equivalence is established by an LLM-based clustering step; \Cref{sec:metric_implementation}). Two generations are considered to agree when their extracted answers are semantically equivalent, i.e., strictly numerically equal (with unit-aware equivalence for SciBench). The Answer Consistency score of generation $i$ is
\begin{equation}
    \mathrm{AC}_i \;=\; \frac{1}{n-1}\sum_{j \neq i} \mathbb{1}\!\left[\, \hat{a}_j \equiv \hat{a}_i \,\right],
\end{equation}
where $\equiv$ denotes semantic equivalence of the extracted answers. As for MCQA, the score depends only on the sampled generations; the gold answer is introduced only during evaluation to assess correctness.

\paragraph{Discrete score values.}
With $n=10$ samples, both scores take values in $\{0.1, 0.2, \dots, 1.0\}$ (and $0.0$ for generations that fail to yield a parseable answer in arithmetic QA). Calibration plots and summary statistics for Answer Consistency accordingly use 11 bins centered on these discrete values (\Cref{sec:metric_implementation}).

\subsubsection{Metric Implementation}
\label{sec:metric_implementation}
\paragraph{Verbalized Uncertainty}
The prompt for Verbalized Uncertainty was adopted directly from the original work \citep{2305.14975}.  
The exact prompt can be found in \path{/llm-uncertainty-bench/seq_ue_calibration/run.py}, where it is provided to the corresponding computation node as \texttt{verbalized\_prompt}.

\paragraph{P(True)}
The prompt used for the P(True) metric follows the formulation of \citep{2207.05221}.  
For APriCoT prompting in MCQA, a minor adaptation was applied, while preserving the core semantics of the original design.  
The exact prompts are available in \path{/llm-uncertainty-bench/seq_ue_calibration/nodes/ptrue.py}.

\paragraph{Answer Consistency}
For this metric, 10 samples were generated per prompt.  
The \emph{Answer Consistency} of a given response is defined as the proportion of semantically equivalent answers within the set of 10 samples.  
Invalid answers are assigned a confidence of 0.0.  
Semantic equivalence was determined according to the task type:
\begin{itemize}
    \item \textbf{Multiple-Choice QA:}  
    Using the APriCoT prompting strategy, each option is independently evaluated, and the model classifies each option as correct or incorrect.  
    Semantic equivalence is established when different generations reach the same classification decision for a given option.
    \item \textbf{Simple Arithmetic QA:}  
    For datasets such as GSM8K, which primarily involve integers and rarely require floating-point precision, the final numeric result was extracted using a dedicated prompt and parsed into a numeric representation.  
    Semantic equivalence is then determined by strict numeric equality of the extracted results.  
    For details on the final answer extraction prompt, please refer to \path{/llm-uncertainty-bench/seq_ue_calibration/nodes/arithmetic_calc.py} in the repository.
    \item \textbf{SciBench:}  
    This dataset presents additional complexity due to intricate computations and the inclusion of physical units, rendering the simple numeric matching used for other arithmetic datasets insufficient.  
    To address this, a specialized clustering prompt was developed to group sampled answers into semantically equivalent categories, with \texttt{Llama-3.3-70B-Instruct} serving as the judging model.  
    Implementation details are provided in  
    \path{/llm-uncertainty-bench/seq_ue_calibration/leaf_nodes/answered_correctly_scibench.py}.
\end{itemize}

For evaluation of the calibration of the Answer Consistency metric, the binning strategy will be slightly modified. Unlike the other methods, this methods yields only discrete confidence values, determined by the number of sampled generations. With 10 generations per question, the resulting confidence scores can only take on values from $0.1$ (indicating that all of the other nine sampled generations resulted in a different answer) to $1.0$ (all generations produced the same result), in steps of $0.1$. For responses that fail to yield a numeric outcome in arithmetic datasets, a confidence score of $0.0$ is assigned. To accommodate these discrete confidence levels, 11 bins centered on the possible certainty scores will be used for generating the calibration plots and summary statistics thereof for the Answer Consistency metric.

\paragraph{Claim-Conditioned Probability (CCP)}
The Claim-Conditioned Probability (CCP) metric, proposed by \citet{2403.04696}, was originally designed for claim-based uncertainty estimation.  
While conceptually valuable, applying CCP to long, complex generations proved challenging: extracting meaningful claims was computationally expensive, often unreliable, and complicated by interdependent claims that hindered aggregation.  
Nevertheless, CCP was included by aggregating token-level confidence scores through multiplicative composition.  
The implementation builds upon the authors' implementation of the metric in their \texttt{LM-Polygraph} framework \citep{fadeeva-etal-2023-lm} and was optimized for improved computational efficiency.

\subsection{Empirical Validation of LLM-as-a-Judge Stability via Model Ensembles}

We acknowledge that the use of LLMs as judges for verifying answer option correctness requires empirical validation to ensure that our evaluation pipeline is not sensitive to the choice of judge model. Motivated by the ensemble-based evaluation strategy of \citet{ielanskyi2025addressing}, we conduct an additional robustness experiment in which we replace a single judge with an ensemble of three diverse models: Ministral-8B-Instruct, Llama-3.3-70B-Instruct, and Qwen3-30B-A3B-Instruct. Using this setup, we re-evaluate a fixed set of 10{,}000 cached responses generated by Llama-3.3-70B on MMLU, ensuring that all judge models operate on identical inputs and evaluation prompts.

Across the ensemble, we observe strong agreement on correctness judgments, with consensus in 9{,}774 out of 10{,}000 cases (97.74\%). This high level of inter-model agreement suggests that the correctness signal used in our benchmark is stable with respect to reasonable variations in judge model selection. Importantly, the small disagreement rate is not systematic and does not indicate any directional bias that would materially affect aggregated results or the comparative conclusions of our study. We therefore conclude that adopting an ensemble-based judging scheme would not meaningfully alter the reported findings.

\subsubsection{Response Lengths}
\label{sec:exp2-response-lengths}
Mean response token lengths across models and datasets are reported in \Cref{tab:token_lengths_combined}.
\begin{table*}[t]
\centering
\small
\setlength{\tabcolsep}{4pt}
\renewcommand{\arraystretch}{1.2}
\begin{tabular}{l c c c c c c c c}
\toprule
\textbf{Model} & \rotatebox{45}{\textbf{MMLU}} & \rotatebox{45}{\textbf{ARC Easy}} & \rotatebox{45}{\textbf{ARC Challenge}} & \rotatebox{45}{\textbf{SciQ}} & \rotatebox{45}{\textbf{GPQA}} & \rotatebox{45}{\textbf{GSM8K}} & \rotatebox{45}{\textbf{SVamp}} & \rotatebox{45}{\textbf{SciBench}} \\
\midrule
gpt-oss-20b & \shortstack[c]{1046 \\{\scriptsize \textcolor{gray}{$\pm$ 431}}} & \shortstack[c]{844 \\{\scriptsize \textcolor{gray}{$\pm$ 283}}} & \shortstack[c]{851 \\{\scriptsize \textcolor{gray}{$\pm$ 305}}} & \shortstack[c]{838 \\{\scriptsize \textcolor{gray}{$\pm$ 273}}} & \shortstack[c]{2838 \\{\scriptsize \textcolor{gray}{$\pm$ 2274}}} & \shortstack[c]{475 \\{\scriptsize \textcolor{gray}{$\pm$ 433}}} & \shortstack[c]{309 \\{\scriptsize \textcolor{gray}{$\pm$ 287}}} & \shortstack[c]{2253 \\{\scriptsize \textcolor{gray}{$\pm$ 2270}}} \\
\midrule
gpt-oss-120b & \shortstack[c]{872 \\{\scriptsize \textcolor{gray}{$\pm$ 323}}} & \shortstack[c]{695 \\{\scriptsize \textcolor{gray}{$\pm$ 305}}} & \shortstack[c]{676 \\{\scriptsize \textcolor{gray}{$\pm$ 313}}} & \shortstack[c]{641 \\{\scriptsize \textcolor{gray}{$\pm$ 257}}} & \shortstack[c]{1754 \\{\scriptsize \textcolor{gray}{$\pm$ 1226}}} & \shortstack[c]{347 \\{\scriptsize \textcolor{gray}{$\pm$ 205}}} & \shortstack[c]{239 \\{\scriptsize \textcolor{gray}{$\pm$ 182}}} & \shortstack[c]{1320 \\{\scriptsize \textcolor{gray}{$\pm$ 964}}} \\
\midrule
Ministral-8B & \shortstack[c]{433 \\{\scriptsize \textcolor{gray}{$\pm$ 173}}} & \shortstack[c]{323 \\{\scriptsize \textcolor{gray}{$\pm$ 102}}} & \shortstack[c]{340 \\{\scriptsize \textcolor{gray}{$\pm$ 111}}} & \shortstack[c]{303 \\{\scriptsize \textcolor{gray}{$\pm$ 98}}} & \shortstack[c]{632 \\{\scriptsize \textcolor{gray}{$\pm$ 196}}} & \shortstack[c]{225 \\{\scriptsize \textcolor{gray}{$\pm$ 91}}} & \shortstack[c]{127 \\{\scriptsize \textcolor{gray}{$\pm$ 128}}} & \shortstack[c]{713 \\{\scriptsize \textcolor{gray}{$\pm$ 234}}} \\
\midrule
Mistral-Nemo-7B & \shortstack[c]{346 \\{\scriptsize \textcolor{gray}{$\pm$ 148}}} & \shortstack[c]{279 \\{\scriptsize \textcolor{gray}{$\pm$ 78}}} & \shortstack[c]{285 \\{\scriptsize \textcolor{gray}{$\pm$ 84}}} & \shortstack[c]{248 \\{\scriptsize \textcolor{gray}{$\pm$ 72}}} & \shortstack[c]{515 \\{\scriptsize \textcolor{gray}{$\pm$ 202}}} & \shortstack[c]{226 \\{\scriptsize \textcolor{gray}{$\pm$ 84}}} & \shortstack[c]{134 \\{\scriptsize \textcolor{gray}{$\pm$ 106}}} & \shortstack[c]{533 \\{\scriptsize \textcolor{gray}{$\pm$ 281}}} \\
\midrule
Mistral-Small-3.2-24B & \shortstack[c]{681 \\{\scriptsize \textcolor{gray}{$\pm$ 444}}} & \shortstack[c]{471 \\{\scriptsize \textcolor{gray}{$\pm$ 130}}} & \shortstack[c]{498 \\{\scriptsize \textcolor{gray}{$\pm$ 169}}} & \shortstack[c]{458 \\{\scriptsize \textcolor{gray}{$\pm$ 131}}} & \shortstack[c]{1337 \\{\scriptsize \textcolor{gray}{$\pm$ 904}}} & \shortstack[c]{435 \\{\scriptsize \textcolor{gray}{$\pm$ 352}}} & \shortstack[c]{284 \\{\scriptsize \textcolor{gray}{$\pm$ 276}}} & \shortstack[c]{1330 \\{\scriptsize \textcolor{gray}{$\pm$ 1033}}} \\
\midrule
Llama-3.3-70B & \shortstack[c]{484 \\{\scriptsize \textcolor{gray}{$\pm$ 324}}} & \shortstack[c]{365 \\{\scriptsize \textcolor{gray}{$\pm$ 83}}} & \shortstack[c]{380 \\{\scriptsize \textcolor{gray}{$\pm$ 93}}} & \shortstack[c]{357 \\{\scriptsize \textcolor{gray}{$\pm$ 77}}} & \shortstack[c]{949 \\{\scriptsize \textcolor{gray}{$\pm$ 536}}} & \shortstack[c]{244 \\{\scriptsize \textcolor{gray}{$\pm$ 74}}} & \shortstack[c]{198 \\{\scriptsize \textcolor{gray}{$\pm$ 291}}} & \shortstack[c]{765 \\{\scriptsize \textcolor{gray}{$\pm$ 384}}} \\
\midrule
DeepSeek-Llama-70B & \shortstack[c]{1290 \\{\scriptsize \textcolor{gray}{$\pm$ 1356}}} & \shortstack[c]{583 \\{\scriptsize \textcolor{gray}{$\pm$ 253}}} & \shortstack[c]{622 \\{\scriptsize \textcolor{gray}{$\pm$ 267}}} & \shortstack[c]{548 \\{\scriptsize \textcolor{gray}{$\pm$ 193}}} & \shortstack[c]{4300 \\{\scriptsize \textcolor{gray}{$\pm$ 2928}}} & \shortstack[c]{597 \\{\scriptsize \textcolor{gray}{$\pm$ 537}}} & \shortstack[c]{331 \\{\scriptsize \textcolor{gray}{$\pm$ 207}}} & \shortstack[c]{4658 \\{\scriptsize \textcolor{gray}{$\pm$ 3398}}} \\
\midrule
Llama-4-Scout-17B & \shortstack[c]{453 \\{\scriptsize \textcolor{gray}{$\pm$ 144}}} & \shortstack[c]{397 \\{\scriptsize \textcolor{gray}{$\pm$ 88}}} & \shortstack[c]{413 \\{\scriptsize \textcolor{gray}{$\pm$ 96}}} & \shortstack[c]{385 \\{\scriptsize \textcolor{gray}{$\pm$ 78}}} & \shortstack[c]{749 \\{\scriptsize \textcolor{gray}{$\pm$ 233}}} & \shortstack[c]{231 \\{\scriptsize \textcolor{gray}{$\pm$ 69}}} & \shortstack[c]{202 \\{\scriptsize \textcolor{gray}{$\pm$ 269}}} & \shortstack[c]{771 \\{\scriptsize \textcolor{gray}{$\pm$ 383}}} \\
\midrule
Qwen3-30B-A3B-Instruct & \shortstack[c]{675 \\{\scriptsize \textcolor{gray}{$\pm$ 432}}} & \shortstack[c]{456 \\{\scriptsize \textcolor{gray}{$\pm$ 140}}} & \shortstack[c]{492 \\{\scriptsize \textcolor{gray}{$\pm$ 168}}} & \shortstack[c]{456 \\{\scriptsize \textcolor{gray}{$\pm$ 197}}} & \shortstack[c]{2824 \\{\scriptsize \textcolor{gray}{$\pm$ 1302}}} & \shortstack[c]{305 \\{\scriptsize \textcolor{gray}{$\pm$ 303}}} & \shortstack[c]{311 \\{\scriptsize \textcolor{gray}{$\pm$ 591}}} & \shortstack[c]{2139 \\{\scriptsize \textcolor{gray}{$\pm$ 1484}}} \\
\midrule
Qwen3-30B-A3B-Thinking & \shortstack[c]{1887 \\{\scriptsize \textcolor{gray}{$\pm$ 1202}}} & \shortstack[c]{1485 \\{\scriptsize \textcolor{gray}{$\pm$ 632}}} & \shortstack[c]{1558 \\{\scriptsize \textcolor{gray}{$\pm$ 692}}} & \shortstack[c]{1496 \\{\scriptsize \textcolor{gray}{$\pm$ 692}}} & \shortstack[c]{6827 \\{\scriptsize \textcolor{gray}{$\pm$ 3213}}} & \shortstack[c]{1095 \\{\scriptsize \textcolor{gray}{$\pm$ 701}}} & \shortstack[c]{801 \\{\scriptsize \textcolor{gray}{$\pm$ 695}}} & \shortstack[c]{6230 \\{\scriptsize \textcolor{gray}{$\pm$ 3252}}} \\
\midrule
DeepSeek-Qwen-32B & \shortstack[c]{1457 \\{\scriptsize \textcolor{gray}{$\pm$ 1605}}} & \shortstack[c]{649 \\{\scriptsize \textcolor{gray}{$\pm$ 274}}} & \shortstack[c]{691 \\{\scriptsize \textcolor{gray}{$\pm$ 310}}} & \shortstack[c]{625 \\{\scriptsize \textcolor{gray}{$\pm$ 253}}} & \shortstack[c]{5251 \\{\scriptsize \textcolor{gray}{$\pm$ 3322}}} & \shortstack[c]{537 \\{\scriptsize \textcolor{gray}{$\pm$ 439}}} & \shortstack[c]{313 \\{\scriptsize \textcolor{gray}{$\pm$ 120}}} & \shortstack[c]{5966 \\{\scriptsize \textcolor{gray}{$\pm$ 3267}}} \\
\midrule
google/gemma-3-27b & \shortstack[c]{426 \\{\scriptsize \textcolor{gray}{$\pm$ 184}}} & \shortstack[c]{287 \\{\scriptsize \textcolor{gray}{$\pm$ 106}}} & \shortstack[c]{299 \\{\scriptsize \textcolor{gray}{$\pm$ 107}}} & \shortstack[c]{280 \\{\scriptsize \textcolor{gray}{$\pm$ 100}}} & \shortstack[c]{856 \\{\scriptsize \textcolor{gray}{$\pm$ 345}}} & \shortstack[c]{248 \\{\scriptsize \textcolor{gray}{$\pm$ 157}}} & \shortstack[c]{180 \\{\scriptsize \textcolor{gray}{$\pm$ 187}}} & \shortstack[c]{951 \\{\scriptsize \textcolor{gray}{$\pm$ 523}}} \\
\midrule
\midrule
Averaged Mean (Instruct) & 500 & 368 & 387 & 355 & 1123 & 273 & 205 & 1029 \\
Averaged Mean (Reasoning) & 1310 & 851 & 880 & 830 & 4194 & 610 & 398 & 4086 \\
Averaged Mean (Total) & 838 & 569 & 592 & 553 & 2403 & 414 & 286 & 2302 \\
\bottomrule
\end{tabular}
\caption{\textbf{Mean response token lengths across models and datasets.} Each cell reports the mean generated response length in tokens (rounded to integers), with the standard deviation shown beneath in smaller gray text. Rows correspond to individual models and columns to evaluation datasets. The final three rows report dataset-wise averages of mean response lengths over instruction-tuned models, reasoning models, and all models, respectively. These aggregates provide a summary of how dataset design, including differences in reasoning demand, affects response length.}
\label{tab:token_lengths_combined}
\end{table*}

\subsubsection{Extensive Calibration Plots}
\label{sec:extensive-calibration-plots-exp2}
In the following, extensive calibration plots for the evaluation of sequence level uncertainty methods are provided (\Cref{fig:metric_PTrue,fig:metric_CCP,fig:metric_Verbalized_Uncertainty,fig:metric_Answer_Consistency}). Again, instruction tuned models are highlighted in blue, while reasoning models are highlighted in green. Darker shading indicates a higher number of items within each confidence bin.

These plots are included in the appendix mainly to visualize systematic patterns in the calibration plots induced by the UQ methods. Full size plots for individual plot assessment are available in the repository.

\clearpage
\onecolumn
\foreach \name/\file in {
    {P(True)}/PTrue,
    {CCP}/CCP,
    {Verbalized Uncertainty}/Verbalized_Uncertainty,
    {Answer Consistency}/Answer_Consistency}{%
    \begin{figure}[H]
        \centering
        \includegraphics[width=\linewidth]{figures/uq_method_\file_calibration_plots.pdf}
        \caption{\textbf{Calibration Plots for \name.} Columns correspond to the models and rows to datasets.}
        \label{fig:metric_\file}
    \end{figure}
}
\clearpage
\twocolumn

\subsubsection{Calibration Correlation by UQ Method}
\label{sec:calibration-correlation}
\Cref{fig:calibration_correlation} shows the per-method distribution of Pearson calibration correlation across all (model, dataset) pairs.
\begin{figure*}[t]
    \centering
     \includegraphics[width=\linewidth]{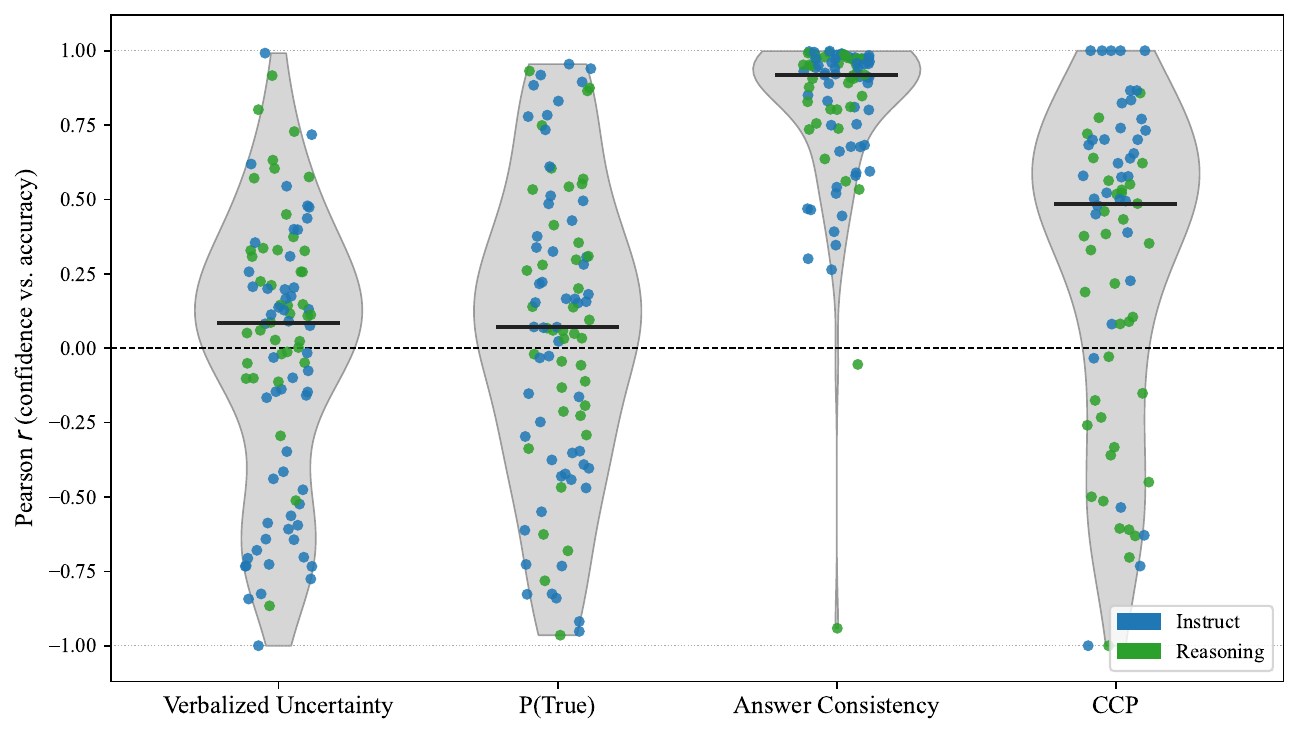}
        \caption[Calibration Correlation by UQ Method]{\textbf{Calibration Correlation by UQ Method.} Each point represents one (model, dataset) pair and shows the Pearson correlation ($r$) between the bin-confidence midpoints and the bucket accuracies of the corresponding calibration curve. Bins with no items are excluded; all non-empty bins contribute equally. A value of $r = 1$ means accuracy increases perfectly in step with confidence; $r = 0$ indicates no linear relationship; negative values indicate miscalibration where higher confidence predicts lower accuracy. Points are coloured by model type (blue = instruct, green = reasoning). The grey violin summarises the distribution across all points for each method; horizontal bars mark the median.}
    \label{fig:calibration_correlation_swarm}
\end{figure*}

\begin{figure*}[t]
    \centering
     \includegraphics[width=\linewidth]{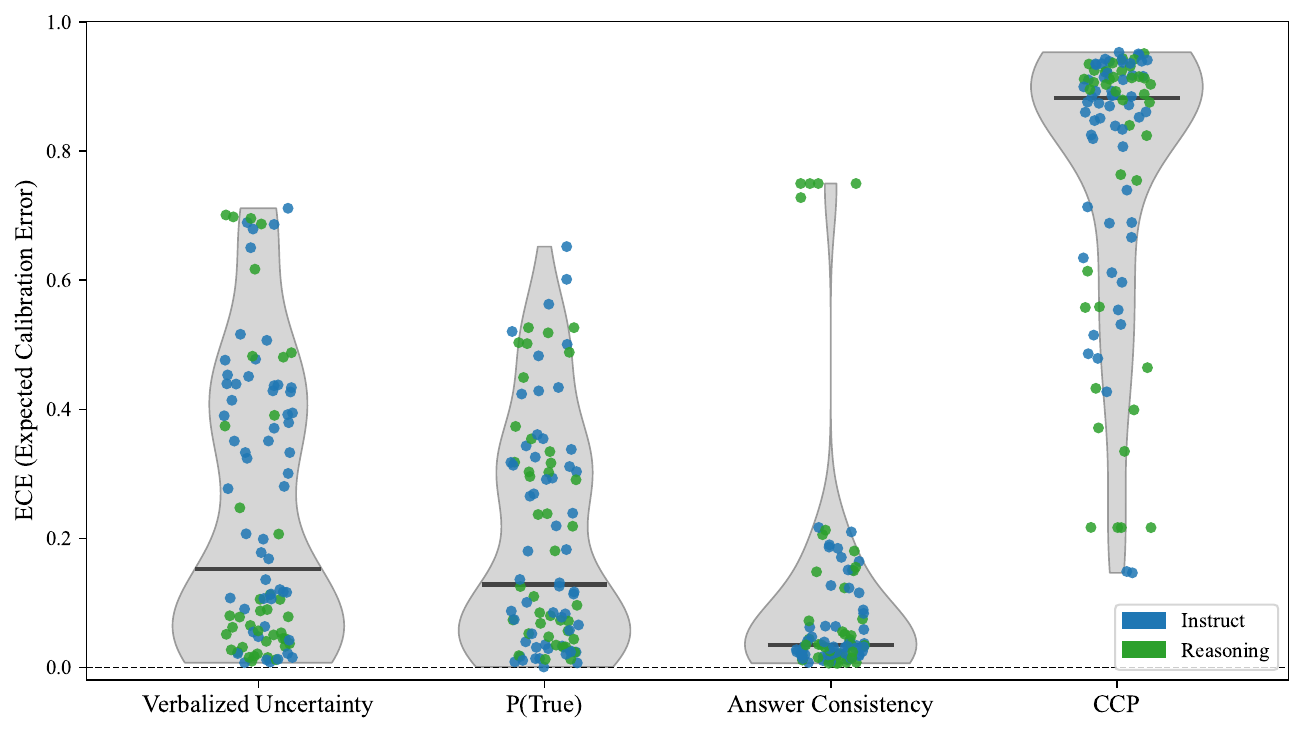}
\caption[Expected Calibration Error by UQ Method]{\textbf{Expected Calibration Error (ECE) by UQ Method.} Distribution of ECE values across all (model, dataset) pairs for each UQ method. Lower values indicate better calibration. Points are coloured by model type (blue = instruct, green = reasoning); violins summarize the overall distribution and horizontal bars indicate the median.}
    \label{fig:calibration_ece_swarm}
\end{figure*}

\begin{figure*}[t]
    \centering
     \includegraphics[width=\linewidth]{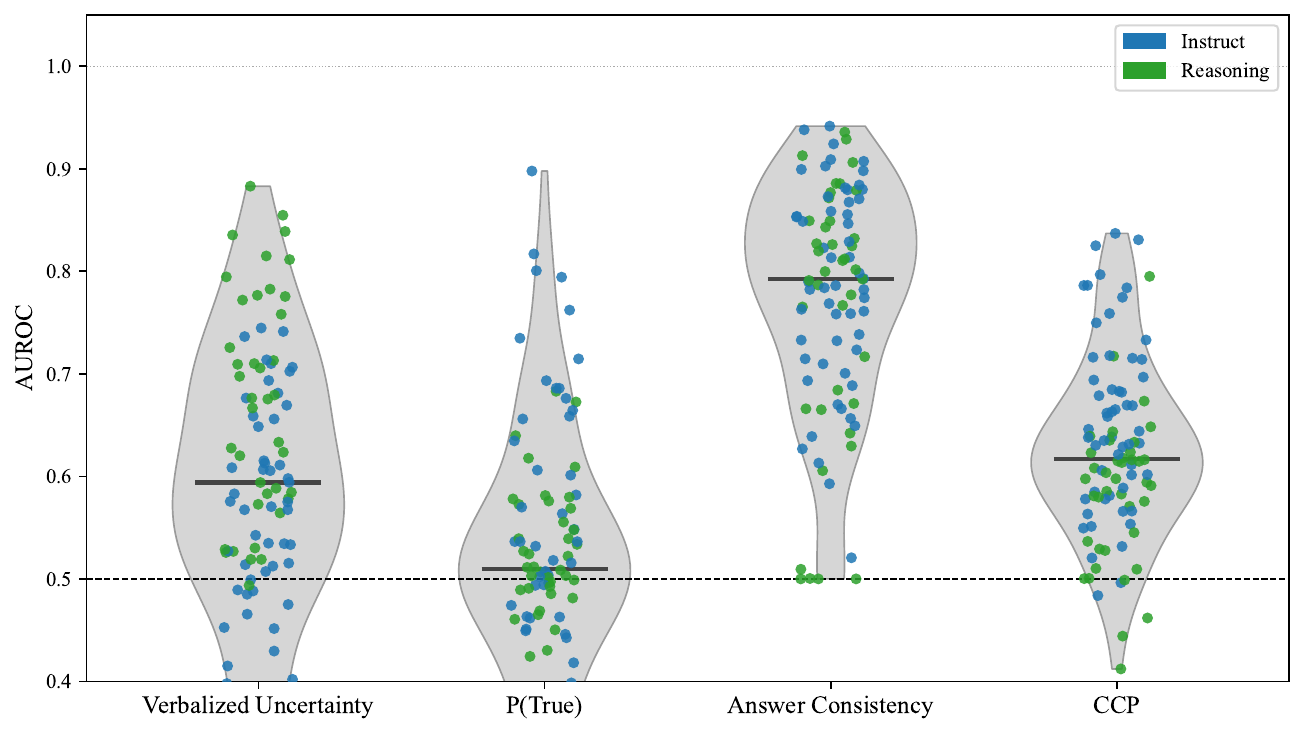}
\caption[AUROC by UQ Method]{\textbf{AUROC by UQ Method.} Distribution of AUROC values across all (model, dataset) pairs for each UQ method. Higher values indicate better discrimination between correct and incorrect responses. Points are coloured by model type (blue = instruct, green = reasoning); violins summarize the overall distribution and horizontal bars indicate the median.}
    \label{fig:calibration_auroc_swarm}
\end{figure*}

\subsubsection{Verbalized Uncertainty Confidence Score Distribution}
\label{sec:verbalized-uncertainty-distribution}
\begin{figure*}[t]
    \centering
     \includegraphics[width=\linewidth]{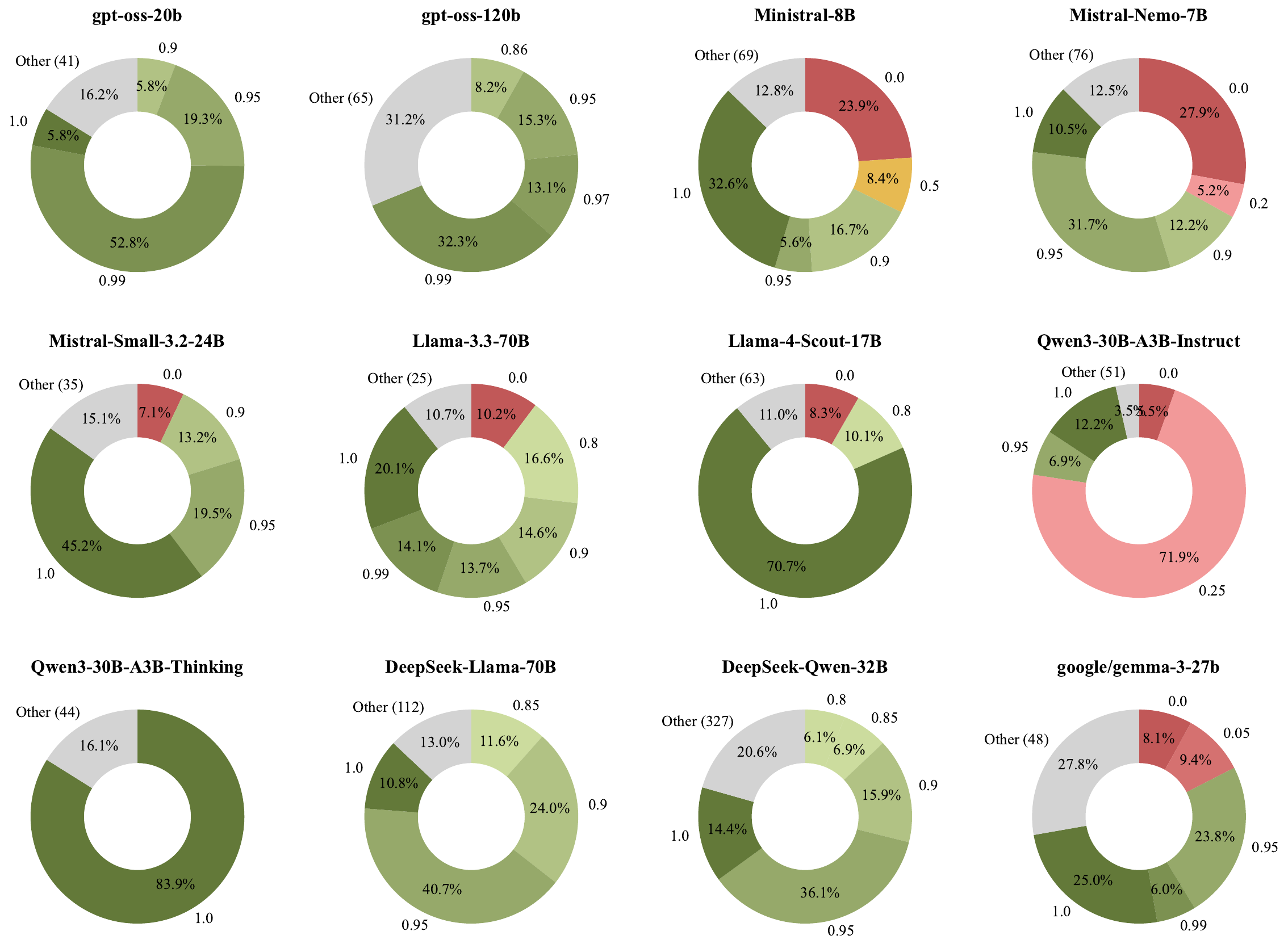}
        \caption[Distribution of Confidence Scores Provided During Verbalized Uncertainty Prompting Across Models]{\textbf{Distribution of Confidence Scores Provided During Verbalized Uncertainty Prompting Across Models.} Value counts are aggregated over all datasets ($57,500$ prompts in total). Confidence scores seen in less than $5\%$ of the total responses have been grouped into ``Other'', with the number of distinct confidence scores shown in brackets.}
    \label{fig:verbalized_value_dist_full}
\end{figure*}

The distribution of confidence scores produced by the Verbalized Uncertainty method across all models is shown in \Cref{fig:verbalized_value_dist_full}.

\subsubsection{P(True) Confidence Score Distribution}
\label{sec:ptrue-distribution}
\begin{figure*}[t]
    \centering
     \includegraphics[width=\linewidth]{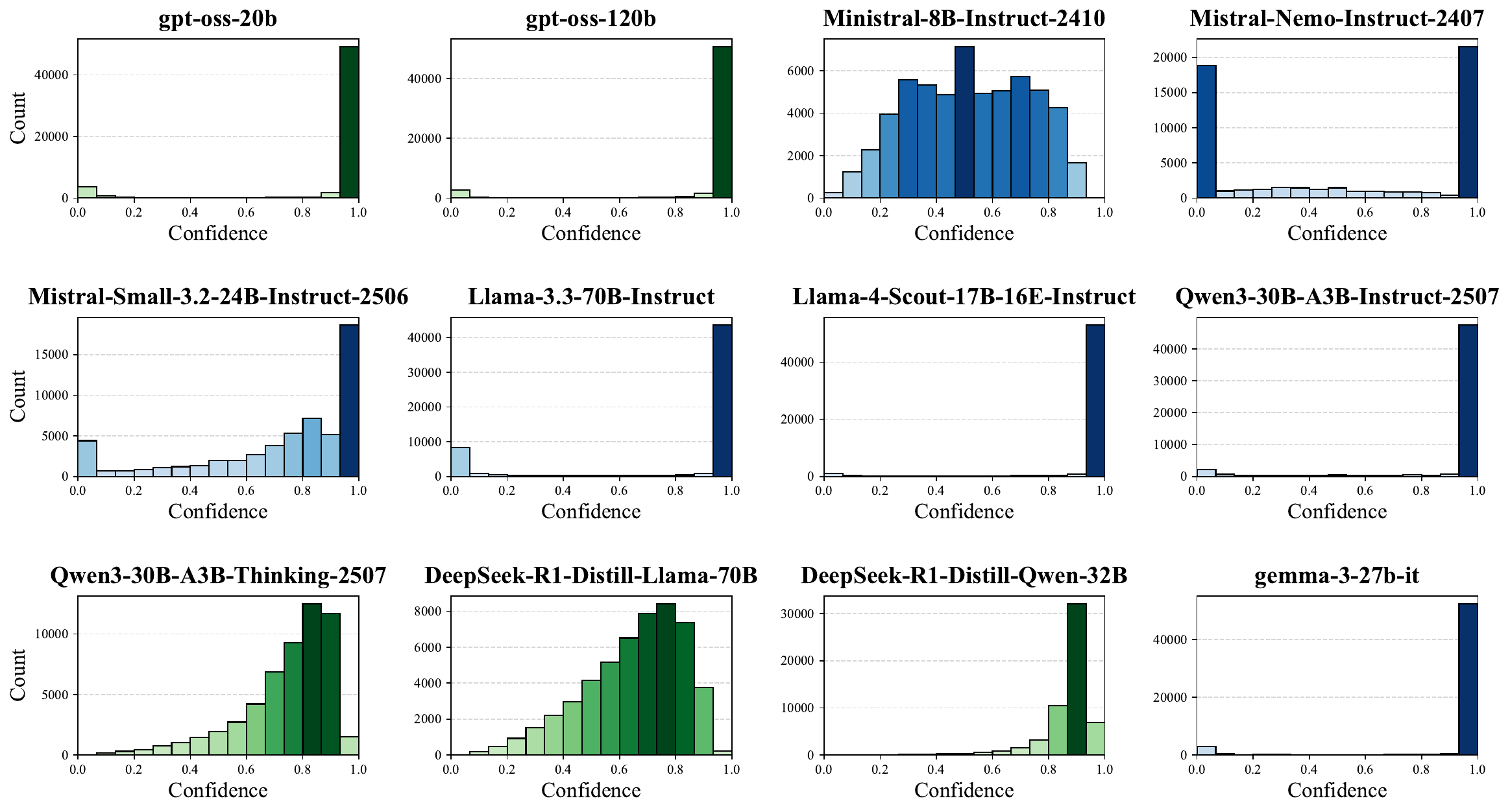}
    \caption[Distribution of Confidence Scores Assigned by P(True)]{%
        \textbf{Distribution of Confidence Scores Assigned by P(True) Across Models.} Confidence scores have been aggregated across all datasets ($57,500$ prompts in total). Most models exhibit a clear polarization towards either \token{(A)} (representing the model's confidence that the answer is true) or \token{(B)} (representing the model's confidence that the answer is false) regarding the token probabilities.}
    \label{fig:ptrue_full}
\end{figure*}

The distribution of confidence scores assigned by P(True) across all models is shown in \Cref{fig:ptrue_full}.

\subsection{Effect of Temperature}

The following sections clarify why additional temperature-sweep experiments would not meaningfully affect our findings or improve the interpretability of the results.

\subsubsection{Experiment 1: Temperature in  Token-Level Calibration}

Experiment 1 evaluates calibration using the label probability of the first generated token under a fixed temperature of 1.0 (reasoning traces, when used, are decoded greedily at the same temperature). The central finding is a near-complete polarization of probability mass onto a single token, driven by extremely large logit margins. This collapse reflects a structural distributional issue, not a mild miscalibration. In such regimes, temperature scaling is ineffective: once margins grow this large, relative logit differences become dominated by noise and cease to carry calibration-relevant information. Adjusting temperature therefore cannot meaningfully “repair” the distribution or alter the calibration behaviour observed.

\subsubsection{Experiment 2: Temperature During Response Generation}

In Experiment 2, temperature affects only the generative content of responses, not the utility of the evaluated UQ methods. Since our goal is not to optimize accuracy or reduce hallucinations, but to identify responses that exhibit uncertain or erroneous reasoning, additional generations at different temperature levels would not strengthen the benchmark. Running a full temperature sweep would require repeating \samplecount{} generations per temperature value, with negligible methodological benefit: temperature primarily alters response variability, not the structure or comparability of the UQ scores themselves.

\subsection{Experiment 2: Temperature for Verbalized Methods}

For verbalized uncertainty methods, we use temperature~1.0 with greedy decoding. This ensures the model provides the reliability judgment it is most confident in. For P(True), the situation mirrors Experiment~1: the method depends on label probabilities, which frequently exhibit polarization and show no reliable alignment with predictive accuracy. Increasing temperature cannot repair this mismatch, nor does it meaningfully alter calibration quality when the underlying distribution is already collapsed.

For these reasons, additional temperature-adjustment experiments would not yield meaningful benefits for either experiment.

\end{document}